\documentclass{article}

\PassOptionsToPackage{numbers, compress}{natbib}

\usepackage[final]{neurips_2024}

\usepackage[utf8]{inputenc} 
\usepackage[T1]{fontenc}    
\usepackage{hyperref}       
\usepackage{url}            
\usepackage{booktabs}       
\usepackage{amsfonts}       
\usepackage{nicefrac}       
\usepackage{microtype}      
\usepackage{xcolor}         
\usepackage{todonotes}
\usepackage{multirow}
\usepackage{graphicx}
\usepackage{amsmath}
\usepackage[capitalize]{cleveref}
\usepackage{xspace}
\usepackage{diagbox}
\usepackage{subfigure}
\usepackage{bbm}
\usepackage{algorithm}
\usepackage{algpseudocode}
\usepackage{authblk}
\usepackage{etoc}

\usepackage{enumitem}
\setlist{leftmargin=8mm}

\usepackage{amsthm}
\DeclareMathOperator*{\argmax}{arg\,max}

\theoremstyle{definition}

\theoremstyle{remark}

\title{Imprecise Label Learning: A Unified Framework for Learning with Various Imprecise Label Configurations}

\author{%
Hao Chen$^{1}$\thanks{haoc3@andrew.cmu.edu},
Ankit Shah$^{1}$,
Jindong Wang$^{2}$,
Ran Tao$^{1}$,
Yidong Wang$^{3}$,
Xiang Li$^{1}$,
\\
Xing Xie$^{2}$,
Masashi Sugiyama$^{4,5}$,
Rita Singh$^{1}$,
Bhiksha Raj$^{1,6}$
}
\affil{\small{$^{1}$Carnegie Mellon University, $^{2}$Microsoft Research,$^{3}$Peking University, 
\\
$^{4}$ RIKEN AIP, $^{5}$The University of Tokyo, $^{6}$MBZUAI
}}

\begin{document}

\maketitle

\begin{abstract}
Learning with reduced labeling standards, such as noisy label, partial label, and supplementary unlabeled data, which we generically refer to as \textit{imprecise} label, is a commonplace challenge in machine learning tasks. 
Previous methods tend to propose specific designs for every emerging imprecise label configuration, which is usually unsustainable when multiple configurations of imprecision coexist. 
In this paper, we introduce imprecise label learning (ILL), a framework for the unification of learning with various imprecise label configurations.
ILL leverages expectation-maximization (EM) for modeling the imprecise label information, treating the precise labels as latent variables.
Instead of approximating the correct labels for training, it considers the entire distribution of all possible labeling entailed by the imprecise information.
We demonstrate that ILL can seamlessly adapt to partial label learning, semi-supervised learning, noisy label learning, and, more importantly, a mixture of these settings, with closed-form learning objectives derived from the unified EM modeling.
Notably, ILL surpasses the existing specified techniques for handling imprecise labels, marking the first practical and unified framework with robust and effective performance across various challenging settings.
We hope our work will inspire further research on this topic, unleashing the full potential of ILL in wider scenarios where precise labels are expensive and complicated to obtain. 
Code is available at: \url{https://github.com/Hhhhhhao/General-Framework-Weak-Supervision}.
\end{abstract}

\section{Introduction}

One of the critical challenges in machine learning is the collection of annotated data for model training \citep{he2016deep, vaswani2017attention, devlin2018bert, dosovitskiy2020image, radford2021learning,openai2023gpt4}. 
Ideally, every data instance would be fully annotated with precise labels. 
However, collecting such data can be expensive, time-consuming, and error-prone. 
Often, the labels can be intrinsically difficult to ascertain precisely. 
Factors such as a lack of annotator expertise and privacy concerns can also negatively affect the quality and completeness of the annotations. 

\begin{figure}[t!]
    \centering
    \hfill
    \subfigure[Full Label]{\label{fig:label_full}\includegraphics[width=0.22\textwidth]{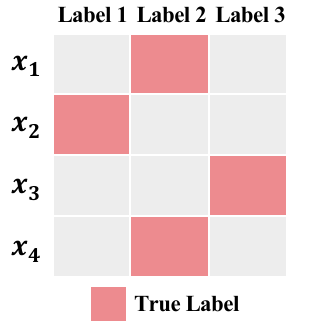}}
    \hfill
    \subfigure[Partial Label]{\label{fig:label_pll}\includegraphics[width=0.22\textwidth]{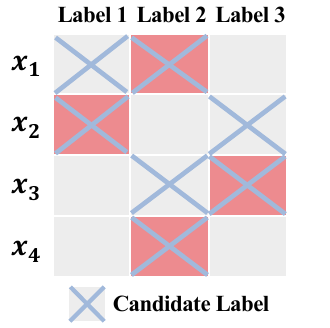}}
    \hfill
    \subfigure[Semi-Supervised]{\label{fig:label_ssl}\includegraphics[width=0.22\textwidth]{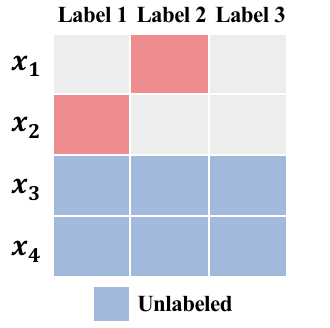}}
    \hfill
    \subfigure[Noisy Label]{\label{fig:label_nll}\includegraphics[width=0.22\textwidth]{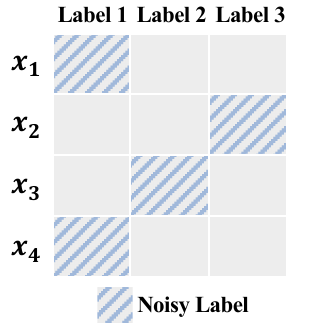}}
    \hfill
    \vspace{-0.1in}
    \caption{Illustration of the full label and imprecise label configurations. We use an example dataset of 4 training instances and 3 classes. (a) Full label, the annotation is a single true label; (b) Partial label, the annotation is a label candidate set containing true label; (c) Semi-supervised, only part of the dataset is labeled, and the others are unlabeled; (d) Noisy label, the annotation is mislabeled.}
    \label{fig:imp-labels}
\vspace{-0.2in}
\end{figure}

In an attempt to circumvent this limitation, several methods have been proposed to permit model learning from the data annotated with reduced labeling standards, which are generally easier to obtain.
We will refer to such labels as \textit{imprecise}. 
\cref{fig:imp-labels} illustrates some typical mechanisms of label imprecision that are commonly addressed in the literature. 
Label imprecision requires a modification of the standard supervised training mechanism to build models for each specific case.
For instance, 
\emph{partial label learning} (PLL) \citep{cour2011learning,Luo2010LearningFC,Feng2020ProvablyCP,Wang2019AdaptiveGG,wen2021leveraged,revisitpllwu22l,wang2022pico} allows instances to have a set of candidate labels, instead of a single definitive one. 
\emph{Semi-supervised Learning} (SSL) \citep{lee2013pseudo,samuli2017temporal,berthelot2019mixmatch,berthelot2019remixmatch,sohn2020fixmatch,xie2020self,zhang2021flexmatch,wang2022debiased,wang2023freematch,chen2023softmatch} seeks to enhance the generalization ability when only a small set of labeled data is available, supplemented by a larger unlabeled set.
\emph{Noisy label learning} (NLL) \citep{Xiao2015LearningFM,alan2016noisyicaspps,Goldberger2016TrainingDN,Ghosh2017RobustLF,Han2018CoteachingRT,Zhang2018GeneralizedCE,Li2018LearningTL,Wang2019SymmetricCE,Liu2020EarlyLearningRP,Li2020DivideMixLW,Ma2020NormalizedLF,wei2022self,Zhang2021LearningNT,wei2023fine} deals with noisy scenarios where the labels are corrupted or incorrect. 
There is a greater variety of other forms of label imprecision, including crowd-sourcing \citep{ibrahim2023deep,wei2023aggregate}, programmable weak supervision \citep{zhang2022survey,wu2022learning}, and bag-level supervision \citep{ilse2018attention,lu2018minimal,scott2020learning,zhang2020aggre,garg2021mixture,feng2021pointwise}, among others. 

While prior arts have demonstrated success in handling individual configurations of label imprecision, their approaches often differ substantially. They are tailored to a \textit{specific} form of imprecision, as depicted in \cref{fig:pipeline}.
Such specificity not only imposes the necessity of devising a solution for emerging types of label imprecision scenarios, but also complicates the deployment in practical settings, where the annotations can be highly complex and may involve \textit{multiple coexisting and interleaved} imprecision configurations.
For instance, considering a scenario where both noisy labels and partial labels appear together, it might be challenging to adapt previous methods in NLL or PLL to this scenario since they either rely on the assumption of definite labels \citep{wei2023aggregate} or the existence of the correct label among label candidates \citep{campagner2021learnability}, thus requiring additional algorithmic design. 
In fact, a few recent works have attempted to address the combinations of imprecise labels in this way, such as partial noisy label \citep{lian2022arnet,xu2023dali} and semi-supervised partial label learning \citep{wang2019partial,wang2020semi}.
However, simply utilizing a more sophisticated or ad-hoc design can hardly scale to other settings. 
In addition, most of these approaches attempt to infer the correct labels given the imprecise information (\textit{e.g.} through consistency with adjacent data \citep{lee2013pseudo,xie2020unsupervised,yao2021instance}, iterative refinement \citep{lv2020progressive,arachie2021constrained}, average over given labels  \citep{hullermeier2015superset,lv2023robustness}, etc., to train the model, which inevitably accumulates error during training.

In this paper, we formulate the problem from a different perspective: rather than taking the imprecise label information provided as a potentially noisy or incomplete attempt at assigning labels to instances, we treat it generically as the information that imposes a deterministic or statistical restriction of the actual applicable true labels.
We then train the model over the distribution of all possible labeling entailed by the given imprecise information.
More specifically, for a dataset with samples $X$ and imprecise label information $I$, we treat the inaccessible full and precise labels $Y$ as a latent variable.
The model is then trained to maximize the likelihood of the provided information $I$.
Since the likelihood computed over the joint probability $P(X, I; \theta) = \sum_{Y}P(X, I, Y; \theta)$ must marginalize out $Y$, the actual information $I$ provided could permit a potentially exponential number of labeling.
To deal with the resulting challenge of maximizing the logarithm of an expectation, we use the common approach of 
\emph{expectation-maximization} (EM) 
\citep{dempster1977maximum}, where the E-step computes the expectation of $P(X, I, Y; \theta)$ given the posterior of current belief $P(Y | X, I; \theta^t)$ at time step $t$ and the M-step maximizes the tight variational lower bound over $P(X, I; \theta)$. 
The overall framework is thus largely agnostic to the various nature of label imprecision, with the imprecise label only affecting the manner in which the posterior $P(Y| X, I; \theta^t)$ is computed. 
In fact, current approaches designed for various imprecise label scenarios can be treated as specific instances of our framework. 
Our approach can serve as a solution towards a \emph{unified and generalized} view for learning with \emph{various} imprecise labels.

While there exist earlier attempts on generalized or EM solutions for different (other) imprecise supervisions or fuzzy observations \citep{denoeux2011maximum,hullermeier2014learning,quost2016clustering,van2017theory,gong2020centroid,zhang2020aggre,chiang2023unified,uumwei23a,xie2024weakly}, they usually require additional assumptions and approximations on the imprecise information for learnablility \citep{campagner2021learnability,campagner2023learning}, thus presenting limited scalability on practical settings \citep{quost2016clustering}.
On the contrary, the unified framework we propose subsumes all of these and naturally extends to the more practical ``mixed'' style of data, where different types of imprecise labels coexist. 
Moreover, for noisy labels, our framework inherently enables the learning of a \textit{noise model}, as we will show in \cref{sec:instantiate}.
Through comprehensive experiments, we demonstrate that the proposed imprecise label learning (ILL) framework 
not only outperforms previous methods for dealing with single imprecise labels of PLL, NLL, and SSL, but also presents robustness and effectiveness for mixed imprecise label learning (MILL) settings, leveraging the full potential to more challenging scenarios.
Our contributions are summarized as follows:
\begin{itemize}
\setlength\itemsep{0em}
    \item We propose an EM framework towards the unification of learning from \emph{various} imprecise labels.
    \item We establish scalable and consistent state-of-the-art (SOTA) performance with the proposed method on partial label learning, semi-supervised learning, and noisy label learning, demonstrating our method's robustness in more diverse, complex label noise scenarios.
    \item To the best of our knowledge, our work is the first to show the robustness and effectiveness of a single unified method for handling the mixture of various imprecise labels. 
\end{itemize}


\section{Preliminary}
\label{sec:preliminary}

\begin{figure}[t!]
    \centering
    \hfill
    \subfigure[Partial Label]{\label{fig:pipe_pll}\includegraphics[width=0.24\textwidth]{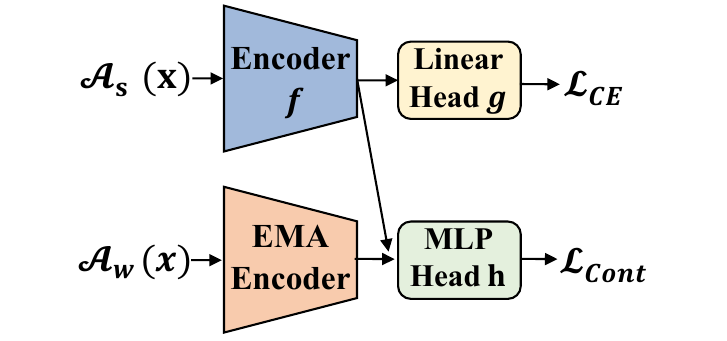}}
    \hfill
    \subfigure[Semi-Supervised]{\label{fig:pipe_ssl}\includegraphics[width=0.24\textwidth]{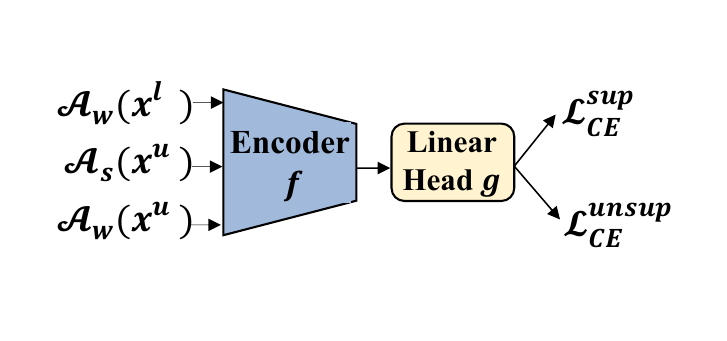}}
    \hfill
    \subfigure[Noisy Label]{\label{fig:pipe_nll}\includegraphics[width=0.24\textwidth]{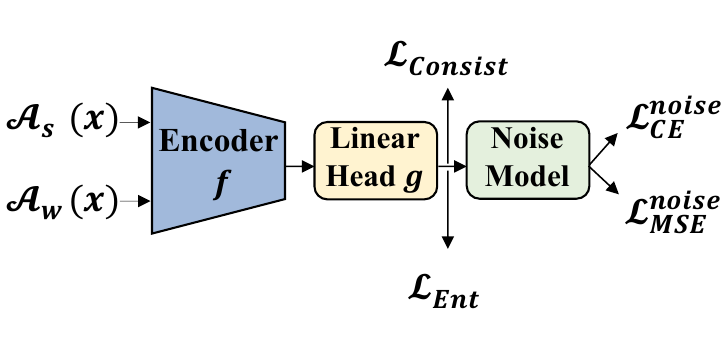}}
    \hfill
    \subfigure[Imprecise Label]{\label{fig:pipe_ill}\includegraphics[width=0.24\textwidth]{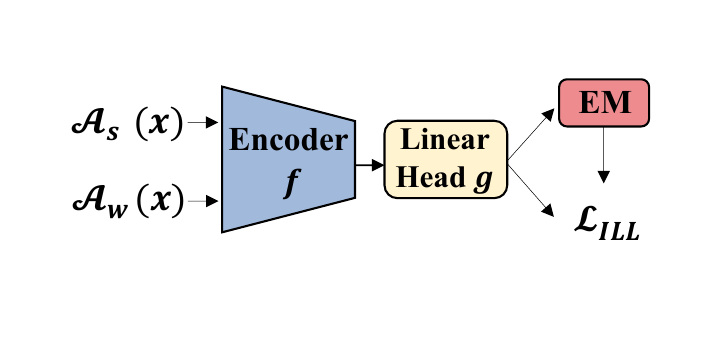}}
    \hfill
    \vspace{-0.1in}
    \caption{Baseline model pipelines for various imprecise label configurations. (a) PiCO \citep{wang2022pico} for partial label learning. (b) FixMatch \citep{sohn2020fixmatch} for semi-supervised learning. (c) SOP \citep{sopliu22w} for noisy label learning. (d) The proposed unified framework. It accommodates \emph{any} imprecise label configurations and also mixed imprecise labels with an EM formulation. 
    }
    \label{fig:pipeline}
\vspace{-0.1in}
\end{figure}

In this section, we illustrate the notations and baselines from different imprecise label settings that adopt various solutions. 
We will show later how our proposed method generalize and subsumes these prior arts.
Let $\mathcal{X}$ denote the input space, and $\mathcal{Y} = [C]:= \{1, \dots, C\}$ represent the label space with $C$ distinct labels.
A fully annotated training dataset of size $N$ is represented as $\mathcal{D} = \{ (\mathbf{x}_i, y_i) \}_{i \in [N]}$. 
Learning with imprecise labels involves approximating the mapping function $f \circ g: \mathcal{X} \rightarrow \mathcal{Y}$ from a training dataset where the true label $y$ is not fully revealed from the annotation process.
Here $f$ is the backbone for feature extraction, $g$ refers to the classifier built on top of the features, and the output from $f \circ g$ is the predicted probability $\mathbf{p}(y|\mathbf{x};\theta)$, where $\theta$ is the learnable parameter for $f \circ g$. 
In this study, we primarily consider three imprecise label configurations (as illustrated in \cref{fig:imp-labels}) and their corresponding representative learning paradigms (as shown in \cref{fig:pipeline}), namely partial label learning, semi-supervised learning, and noisy label learning. 

\textbf{Partial label learning (PLL)}. 
PLL aims to learn with a candidate label set $\mathbf{s} \subset \mathcal{Y}$, where the ground truth label $y \in \mathcal{Y}$ is concealed in $\mathbf{s} $. 
The training data for partial labels thus becomes $\mathcal{D}_{\mathrm{PLL}} = \{ (\mathbf{x}_i, \mathbf{s} _i)\}_{i \in [N]}$. 
PiCO \citep{wang2022pico} is a recent contrastive method that employs class prototypes to enhance label disambiguation (as shown in \cref{fig:pipe_pll}). 
It optimizes the cross-entropy (CE)\footnote{For simplicity, we use $\mathcal{L}_{\mathbf{CE}}$ for labels of the formats of class indices, one-hot vectors, and class probabilities.} loss between the prediction of the augmented training sample $\mathcal{A}_{\mathrm{w}}(\mathbf{x})$ and the disambiguated labels $\hat{\mathbf{s}}$. 
PiCO learns a set of class prototypes from the features associated with the same pseudo-targets. 
A contrastive loss, based on MOCO \citep{moco2020}, is employed to better learn the feature space,
drawing the projected and normalized features $\mathbf{z}_\mathrm{w}$ and $\mathbf{z}_\mathrm{s}$ 
of the two augmented versions of data $\mathcal{A}_\mathrm{w}(\mathbf{x})$ and $\mathcal{A}_\mathrm{s}(\mathbf{x})$ \footnote{We use $\mathcal{A}_\mathrm{w}$ to indicate the weaker data augmentation and $\mathcal{A}_\mathrm{s}$ to indicate the stronger data augmentation.} closer.
The objective of PiCO is formulated as:
\begin{equation}
    \mathcal{L}_{\mathrm{PiCO}} = \mathcal{L}_{\mathrm{CE}} \left(\mathbf{p}(y|\mathcal{A}_\mathrm{w}(\mathbf{x});\theta), \hat{\mathbf{s}} \right) + \mathcal{L}_{\mathrm{Cont}} \left( \mathbf{z}_\mathrm{w}, \mathbf{z}_\mathrm{s}, \mathcal{M} \right).
\label{eq:pico}
\end{equation}

\textbf{Semi-supervised learning (SSL)}. 
For SSL, we can define the labeled dataset as $\mathcal{D}_{\mathrm{SSL}}^\mathrm{L} = \{(\mathbf{x}^\mathrm{l}_i, y^\mathrm{l}_i)\}_{i \in [N^\mathrm{L}]}$, and the unlabeled dataset as $\mathcal{D}^\mathrm{U} = \{ \mathbf{x}^\mathrm{u}_j \}_{j \in [N^\mathrm{L} + 1, N^\mathrm{L}+N^\mathrm{U}]}$, with $N^\mathrm{L} \ll N^\mathrm{U}$.
A general confidence-thresholding based self-training \citep{xie2020unsupervised,sohn2020fixmatch} pipeline for SSL is shown in \cref{fig:pipe_ssl}.
Consider FixMatch \citep{sohn2020fixmatch} as an example; there are usually two loss components: the supervised CE loss on labeled data and the unsupervised CE loss on unlabeled data. 
For the unsupervised objective, 
the pseudo-labels $\hat{y}^\mathrm{u}$ from the network itself are used to train on the unlabeled data. 
A ``strong-weak'' augmentation \citep{xie2020unsupervised} is commonly adopted.
To ensure the quality of the pseudo-labels, only the pseudo-labels whose confidence scores $\hat{p}^\mathrm{u}$ are greater than a threshold $\tau$ are selected to participate in training:
\begin{equation}
    \mathcal{L}_{\mathrm{Fix}} = \mathcal{L}_{\mathrm{CE}} \left( \mathbf{p}(y|\mathcal{A}_\mathrm{w}(\mathbf{x}^\mathrm{l});\theta), y^\mathrm{l} \right) + \mathbbm{1}\left( \hat{p}^\mathrm{u} \geq \tau \right) \mathcal{L}_{\mathrm{CE}} \left( \mathbf{p}(y|\mathcal{A}_\mathrm{s}(\mathbf{x}^\mathrm{u});\theta)  , \hat{y}^\mathrm{u} \right).
\label{eq:fixmatch}
\end{equation}

\textbf{Noisy label learning (NLL)}. 
NLL aims at learning with a dataset of corrupted labels, $\mathcal{D}_{\mathrm{NLL}} = \{ (\mathbf{x}_i, \hat{y}_i) \}_{i \in [N]}$.
We illustrate the NLL pipeline (in \cref{fig:pipe_nll}) with the recent sparse over-parameterization (SOP) model \citep{sopliu22w}, where a sparse \textit{noise model} consisting of parameters $\mathbf{u}_i, \mathbf{v}_i \in [-1, 1]^C$ for each sample is adopted. 
The noise model transforms the network prediction from the true label distribution into the noisy label distribution.
A CE loss and a mean-squared-error (MSE) loss optimize parameter $\{\mathbf{u}_i\}$ and $\{\mathbf{v}_i\}$ respectively:
\begin{equation}
    \mathcal{L}_{\mathrm{SOP}} = \mathcal{L}_{\mathrm{CE}} \left( \phi \left( \mathbf{p}(y|\mathcal{A}_\mathrm{w}(\mathbf{x});\theta) + \mathbf{m} \right), \hat{y} \right) + \mathcal{L}_{\mathrm{MSE}} \left( \mathbf{p}(y|\mathcal{A}_\mathrm{w}(\mathbf{x});\theta) + \mathbf{m}, \hat{y} \right),
\end{equation}
where $\phi$ denotes the $L_\infty$ normalization and $\mathbf{m}_i = \mathbf{u}_i \odot \mathbf{u}_i \odot \hat{\mathbf{y}}^{\mathrm{oh}}_i - \mathbf{v}_i \odot \mathbf{v}_i\odot \left(1- \hat{\mathbf{y}}^{\mathrm{oh}}_i \right)$, with $\hat{\mathbf{y}}^{\mathrm{oh}}_i$ referring to the one-hot version of $y_i$. Consistency regularization with strong-weak augmentation and entropy class-balance regularization are additionally utilized for better performance in SOP \citep{sopliu22w}.


\section{Imprecise Label Learning}

Although current techniques demonstrate potential in addressing particular forms of imprecise labels, they frequently fall short in adaptability and transferability to more complicated and more realistic scenarios where multiple imprecise label types coexist and interleave. This section first defines the proposed expectation-maximization (EM) formulation for learning with various imprecise labels. Then, we demonstrate that our unified framework seamlessly extends to partial label learning, semi-supervised label learning, noisy label learning, and the more challenging setting of mixed imprecise label learning.
Connections and generalization to previous pipelines can also be drawn clearly under the proposed EM framework.

\subsection{A Unified Framework for Learning with Imprecise Labels}

\textbf{Exploiting information from imprecise labels}.
The challenge of learning with imprecise labels lies in learning effectively with inaccurate or incomplete annotation information. 
Per the analysis above, prior works catering to specific individual imprecise labels either explicitly or implicitly attempt to infer the precise labels from the imprecise label information. 
For example, partial label learning concentrates on the disambiguation of the ground truth label from the label candidates \citep{wang2022pico,lian2022irnet,xu2023dali} or averaging equally over the label candidates \citep{hullermeier2006learning}. 
In semi-supervised learning, after the model initially learns from the labeled data, the pseudo-labels are treated as correct labels and utilized to conduct self-training on the unlabeled data \citep{arazo2020pseudo,sohn2020fixmatch}.
Similarly, for noisy label learning,  an integral part that helps mitigate overfitting to random noise is the implementation of an accurate noise model capable of identifying and rectifying the incorrect labels \citep{Li2020DivideMixLW, sopliu22w}, thereby ensuring the reliability of the learning process.
However, inferring the correct labels from the imprecise labels or utilizing the imprecise labels directly can be very challenging and usually leads to errors accumulated during training \citep{arazo2020pseudo,chen2022debiased}, which is also known as the confirmation bias. 
In this work, we take a different approach: we consider all possible labeling along with their likelihood that the imprecise labels fulfill to train the model, rather than using a single rectified label from the imprecise information. 
Such an approach also eliminates the requirements for designing different methods for various imprecise labels and provides a unified formulation instead, where closed-form solutions can be derived.

\textbf{A unified framework for learning with imprecise labels (ILL)}.
Let $\{\mathbf{x}_i\}_{i \in [N]}$ represent the features as realizations from $X$ and $\{y_i\}_{i \in [N]}$ represent their precise labels as realizations from $Y$ for the training data. Ideally, $Y$ would be fully specified for $X$. In the imprecise label scenario, however, $Y$ is not provided; instead we obtain imprecise label information $I$.  
We view $I$ not as \textit{labels}, but more abstractly as a variable representing the \textit{information} about the labels. From this perspective, the actual labels $Y$ would have a distribution $P(Y|I)$, and $I$ can present in various forms.
When the information $I$ provided is the precise true label of the data,  $P(Y|I)$ would be a delta distribution, taking a value $1$ at the true label, and $0$ elsewhere. If $I$ represents partial labels, then $P(Y|I)$ would have non-zero value over the candidate labels, and be $0$ elsewhere. 
When $I$ represents a set of noisy labels, $P(Y|I)$ would represent the distribution of the true labels, given the noisy labels.
When $I$ does not contain any information, i.e., unlabeled data, $Y$ can take any value. 

By the maximum likelihood estimation (MLE) principle, we must estimate the model to maximize the likelihood of the data/information we have been provided, namely $X$ and $I$. Let $P(X, I; \theta)$ represent a parametric form for the joint distribution of $X$ and $I$\footnote{The actual parameters $\theta$ may apply only to some component such as $P(Y|X; \theta)$ of the overall distribution; we will nonetheless tag the entire distribution $P(X,I; \theta)$ with $\theta$ to indicate that it is dependent on $\theta$ overall.} Explicitly considering the labels $Y$, we have $ P(X, I; \theta) = \sum_Y P(X, Y, I; \theta)$.
The maximum likelihood principle requires us to find:
\begin{equation}
\theta^*  = \argmax_\theta~ \log P(X, I; \theta)  = \argmax_\theta~ \log \sum_Y P(X, Y, I; \theta),
\label{eq:basic}
\end{equation}
with $\theta^*$ denotes the optimal value of $\theta$.
\cref{eq:basic} features the log of an expectation and cannot generally be solved in closed-form, and requires iterative hill-climbing solutions. Of these, arguably the most popular is the expectation-maximization (EM) algorithm \citep{dempster1977maximum}, which iteratively maximizes a tight variational lower bound on the log-likelihood. 
In our case, applying it becomes:
\begin{equation}
\begin{split}
    \theta^{t+1} &= \argmax_{\theta} \mathbb{E}_{Y|X, I;\theta^t} \left[ \log P(X, Y, I; \theta) \right] \\ 
    &= \argmax_{\theta} \mathbb{E}_{Y|X, I;\theta^t} \left[ \log P(Y | X; \theta) + \log P(I|X,Y;\theta) \right],
\end{split}
\label{eq:em}
\end{equation}
where $\theta^t$ is the $t^{\rm th}$ estimate of the optimal $\theta$. 
Note that $P(X;\theta)$ is omitted from \cref{eq:em} since $P(X)$ does not rely on $\theta$.
The detailed derivation of the variational lower bound is shown in \cref{sec:append-var-lower-bound}.
There are several implications from \cref{eq:em}. 
(i) The expectation over the posterior $P(Y|X,I;\theta^t)$ equates to considering \textit{all} labeling entailed by the imprecise label information $I$,  rather than any single (possibly corrected) choice of label. 
For independent instances setting mostly studied in this paper, we can derive closed-form training objectives from this formulation as shown in \cref{sec:instantiate}.
(ii) The property of the second term $\log P(I|X, Y; \theta)$ is dependent on the nature of imprecise label $I$. 
If $I$ is derivable from true labels $Y$, such as the actual labels or the label candidates, it can be reduced to $P(I|Y)$, \textit{i.e.}, the probability of $I$ is no longer dependent on $X$ or $\theta$ and thus can be ignored from \cref{eq:em}. 
If $I$ represents the noisy labels, $P(I|X,Y;\theta)$ instead includes a potentially learnable noise model.  
(iii) It is a general framework towards the unification of any label configuration, including full labels, partial labels, low-resource labels, noisy labels, etc.
In this work, we specialize the proposed EM framework to PLL, SSL, NLL, and the mixture of them in the following.

\subsection{Instantiating the Unified EM Formulation}
\label{sec:instantiate}


We illustrate how to seamlessly expand the formulation from \cref{eq:em} to partial label learning, semi-supervised learning, noisy label learning, and mixture settings, with derived closed-form loss function\footnote{To formulate the loss function, we convert the problem to minimization of the negative log-likelihood.} for each setting here. 
The actual imprecise labels only affect the manner in which the posterior $P(Y| X, I; \theta^t)$ is computed for each setting.  
We show that all learning objectives derived from \cref{eq:em} naturally include a consistency term with the posterior as the soft target.
We also demonstrate that the proposed unified EM framework closely connects with the prior arts, which reveals the potential reason behind the success of these techniques.
Note that while we only demonstrate the application of the proposed framework to four settings here, it can also be flexibly extended to other settings. 
More details of derivation below are shown in \cref{sec:appendix-method}.

\textbf{Partial label learning (PLL)}. 
The imprecise label $I$ for partial labels is defined as the label candidate sets $S$ containing the true labels.
These partial labels indicate that the posterior $P(Y|X,S;\theta^t)$ can only assign its masses on the candidate labels.
Since $S$ can be derived from true labels $Y$, $P(S|X, Y;\theta)$ reduces to $P(S|Y)$, and thus can be ignored.
We also demonstrate with instance dependent partial labels that maintains $P(S|X, Y;\theta)$ in \cref{sec:append-exp-rcr}.
Defining the label candidates as $\{\mathbf{s}_i\}_{i \in [N]}$ and substituting it in \cref{eq:em}, we have the loss function of PLL derived using ILL framework:
\begin{equation}
\begin{split}
    \mathcal{L}_{\mathrm{ILL}}^{\mathrm{PLL}} 
    = - \sum_{Y \in [C]} P(Y|X, S;\theta^t) \log P(Y | X; \theta) \equiv \mathcal{L}_{\mathrm{CE}} \left( \mathbf{p}(y|\mathcal{A}_\mathrm{s}(\mathbf{x});\theta),  \mathbf{p}(y|\mathcal{A}_\mathrm{w}(\mathbf{x}), \mathbf{s};\theta^t) \right),
\end{split}
\label{eq:ill-pll}
\end{equation}
where $\mathbf{p}(y|\mathcal{A}_\mathrm{w}(\mathbf{x}), \mathbf{s};\theta^t)$ is the normalized probability that $\sum_{k \in C} p_{k} = 1$, and $p_{k} = 0, \forall k \in \mathbf{s}$. 
\cref{eq:ill-pll} corresponds exactly to consistency regularization \citep{xie2020unsupervised}, with the normalized predicted probability as the soft pseudo-targets. 
We use $\mathcal{A}_s$ and $\mathcal{A}_w$ to denote the strong and weak augmentation as stated earlier.
This realization on PLL shares similar insights as \citep{revisitpllwu22l} which exploits a gradually induced loss weight for PLL on multiple augmentations of the data. 
However, our framework is much simpler and more concise as shown in \cref{sec:append-exp-rcr}, which does not require additional techniques.

\textbf{Semi-supervised learning (SSL)}
In SSL, the input $X$ consists of the labeled data $X^\mathrm{L}$ and the unlabeled data $X^\mathrm{U}$. 
The imprecise label for SSL is realized as the limited number of full labels $Y^\mathrm{L}$ for $X^\mathrm{L}$. 
The labels $Y^\mathrm{U}$ for unlabeled $X^\mathrm{U}$ are unknown and become the latent variable. 
Interestingly, for the unlabeled data, there is no constraint on possible labels it can take. 
The posterior $P(Y^\mathrm{U}|X^{\mathrm{L}}, X^{\mathrm{U}}, Y^{\mathrm{L}}; \theta)$, which is the actual prediction from the network, can be directly utilized as soft targets for self-training. 
Since $Y^\mathrm{L}$ is conditionally independent with $Y^\mathrm{U}$ given $X$, the second term of \cref{eq:em}: $P(Y^\mathrm{L}|X^\mathrm{L}, X^\mathrm{U}, Y^\mathrm{U};\theta)$, is reduced to $P(Y^\mathrm{L}|X^\mathrm{L};\theta)$, which corresponds to the supervised objective on labeled data.
The loss function for SSL thus becomes:
\begin{equation}
\begin{split}
    \mathcal{L}_{\mathrm{ILL}}^{\mathrm{SSL}} &= - \sum_{Y \in [C]} P(Y^\mathrm{U} | X^\mathrm{U}, X^\mathrm{L}, Y^\mathrm{L};\theta^t) \log P(Y^\mathrm{U} | X^\mathrm{U}, X^\mathrm{L};\theta) - \log P(Y^\mathrm{L} | X^\mathrm{L}; \theta)  \\
    &\equiv  \mathcal{L}_{\mathrm{CE}} \left( \mathbf{p}(y|\mathcal{A}_\mathrm{s}(\mathbf{x}^\mathrm{u});\theta)  , \mathbf{p}(y|\mathcal{A}_\mathrm{w}(\mathbf{x}^\mathrm{u});\theta^t)  \right) + \mathcal{L}_{\mathrm{CE}} \left( \mathbf{p}(y|\mathcal{A}_\mathrm{w}(\mathbf{x}^\mathrm{l});\theta), y^\mathrm{l} \right)
\end{split}
\label{eq:ill-ssl}
\end{equation}
The first term corresponds to the unsupervised consistency regularization usually employed in SSL, and the second term refers to the supervised CE loss only on labeled data.
\cref{eq:ill-ssl} has several advantages over the previous methods. 
It adopts the prediction as soft-targets of all possible labeling on unlabeled data, potentially circumventing the confirmation bias caused by pseudo-labeling and naturally utilizing all unlabeled data which resolves the quantity-quality trade-off commonly existing in SSL \citep{sohn2020fixmatch,chen2023softmatch}.
It also indicates that previous pseudo-labeling with confidence threshold implicitly conducts the EM optimization, where the maximal probable prediction approximates the expectation, and the degree of the approximation is determined by the threshold $\tau$, rationalizing the effectiveness of dynamic thresholding.

\textbf{Noisy label learning (NLL)}.
Things become more complicated here since the noisy labels $\hat{Y}$ do not directly reveal the true information about $Y$, thus $P(\hat{Y}|Y, X;\theta)$ inherently involves a noise model that needs to be learned.  
We define a simplified instance-independent\footnote{A more complicated instance-dependent noise model $\mathcal{T}(\hat{Y}|Y, X;\omega)$ can also be formulated under our unified framework, but not considered in this work. Also, since we use $\mathcal{T}$ both in forward fashion and backward fashion, it is unidentifiable in this work.} noise transition model $\mathcal{T}(\hat{Y} | Y;\omega)$ with parameters $\omega$, and take a slightly different way to formulate the loss function for NLL from the ILL framework:
\begin{equation}
\begin{split}
    \mathcal{L}_{\mathrm{ILL}}^{\mathrm{NLL}} &= - \sum_{Y \in [C]} P(Y | X, \hat{Y}; \theta^t,  \omega^t) \log P(Y | X, \hat{Y}; \theta,  \omega^t) - \log P(\hat{Y} | X; \theta, \omega) \\ 
    &\equiv \mathcal{L}_{\mathrm{CE}}\left( \mathbf{p}(y|\mathcal{A}_\mathrm{s}(\mathbf{x}), \hat{y};\theta, \omega^t), 
    \mathbf{p}(y| \mathcal{A}_\mathrm{w}(\mathbf{x}), \hat{y}; \theta^t, \omega^t) \right) 
    + \mathcal{L}_{\mathrm{CE}} \left( \mathbf{p}(\hat{y}|\mathcal{A}_\mathrm{w}(\mathbf{x});\theta,\omega), \hat{y} \right),
\end{split}
\end{equation}
where the parameters $\omega$ and $\theta$ are learned end-to-end. 
The first term corresponds to the consistency regularization of prediction conditioned on noisy labels and the second term corresponds to the supervised loss on noisy predictions that are converted from the ground truth predictions.
Both quantities are computed using the noise transition model given the noisy label $\hat{y}$:
\begin{equation}
    \mathbf{p}(y|\mathbf{x}, \hat{y};\theta, \omega^t) \propto \mathbf{p}(y|\mathbf{x};\theta) \mathcal{T}(\hat{y} | y; \omega^t) , \text{and }
    \mathbf{p}(\hat{y}|\mathbf{x};\theta,\omega) = \sum_{y \in [C]} \mathbf{p}(y|\mathbf{x};\theta)  \mathcal{T}(\hat{y} | y; \omega).
\label{eq:ill-nll}
\end{equation}

\textbf{Mixture imprecise label learning (MILL)}. 
We additionally consider a more practical setting, mixture of imprecise label learning, with partial labels, noisy labels, and unlabeled data interleaved together. 
On the unlabeled data, the unsupervised objective is the same as the unsupervised consistency regularization of SSL as shown in \cref{eq:ill-ssl}. 
The labeled data here present partial and noisy labels $\hat{\mathbf{s}}$.
Thus the noisy supervised objective in \cref{eq:ill-nll} becomes the supervised consistency regularization as in \cref{eq:ill-pll} of partial label setting to train the noise transition model, and the noisy unsupervised objective becomes the consistency regularization of the prediction conditioned on noisy partial labels. 
Thus we have the loss function for MILL derived as:
\begin{equation}
\begin{split}
    \mathcal{L}_{\mathrm{ILL}}^{\mathrm{MILL}} &=  \mathcal{L}_{\mathrm{CE}}\left(\mathbf{p}\left(y \mid \mathcal{A}_{\mathrm{s}}(\mathbf{x}^l), \hat{\mathbf{s}}^l ; \theta, \omega^t\right), \mathbf{p}\left(y \mid \mathcal{A}_{\mathrm{w}}(\mathbf{x}^l), \hat{\mathbf{s}}^l ; \theta^t, \omega^t\right)\right) \\
    &+ \mathcal{L}_{\mathrm{CE}}\left(\mathbf{p}\left(\hat{y} \mid \mathcal{A}_{\mathrm{w}}(\mathbf{x}^l) ; \theta, \omega\right), \hat{\mathbf{s}}^l\right) \\
    &+ \mathcal{L}_{\mathrm{CE}} \left( \mathbf{p}(y|\mathcal{A}_\mathrm{s}(\mathbf{x}^\mathrm{u});\theta)  , \mathbf{p}(y|\mathcal{A}_\mathrm{w}(\mathbf{x}^\mathrm{u});\theta^t)  \right)
\end{split}
\end{equation}

We can compute both quantity through the noise transition model:
\begin{equation}
    \mathbf{p}(y|\mathbf{x}, \hat{\mathbf{s}};\theta, \omega^t) \propto \mathbf{p}(y|\mathbf{x};\theta) \prod_{\hat{y} \in \hat{\mathbf{s}}} \mathcal{T}(y | \hat{y}; \omega^t) , \text{and }
    \mathbf{p}(\hat{y} |\mathbf{x};\theta,\omega) = \sum_{y \in [C]} \mathbf{p}(y|\mathbf{x};\theta)  \mathcal{T}(\hat{y} | y; \omega).
\end{equation}

\section{Experiments}

In this section, we conduct extensive experiments to evaluate ILL. 
Albeit simple, the ILL framework achieves comparable state-of-the-art performance regarding previous methods on partial label learning, semi-supervised learning, and noisy label learning. 
Moreover, our experiments show that ILL could be easily extended to a more practical setting with a mixture of various imprecise label configurations. 
For all settings, we additionally adopt an entropy loss for balancing learned cluster sizes \citep{bridle1991unsup,joulin2012convex}, similarly as \citep{sopliu22w, wang2023freematch}. 
Experiments are conducted with three runs using NVIDIA V100 GPUs.

\subsection{Partial Label Learning}

\textbf{Setup}. 
Following \citep{wang2022pico}, we evaluate our method on partial label learning setting using CIFAR-10 \citep{krizhevsky2009learning}, CIFAR-100 \citep{krizhevsky2009learning}, and CUB-200 \citep{welinder2010caltech}.
We generate partially labeled datasets by flipping negative labels to false positive labels with a probability $q$, denoted as a partial ratio. 
The $C - 1$ negative labels are then uniformly aggregated into the ground truth label to form a set of label candidates. 
We consider $q \in \{0.1, 0.3, 0.5\}$ for CIFAR-10, $q \in \{0.01, 0.05, 0.1\}$ for CIFAR-100, and $q=0.05$ for CUB-200. 
We choose six baselines for PLL using ResNet-18 \citep{he2016deep}: LWS \citep{wen2021leveraged}, PRODEN \citep{lv2020progressive}, CC \citep{Feng2020ProvablyCP}, MSE and EXP \citep{feng2020learning}, and PiCO~\citep{wang2022pico}. The detailed hyper-parameters, comparison with the more recent method R-CR \citep{revisitpllwu22l} that utilizes a different training recipe and model \citep{zagoruyko2016wide}, and comparison with instance-dependent partial labels \citep{xu2021instance} are shown in \cref{sec:append-exp-rcr}.

\begin{table}[t!]
    \centering
    \caption{Accuracy of different partial ratio $q$ on CIFAR-10, CIFAR-100, and CUB-200 for \textbf{partial label learning}. 
    The best and the second best results are indicated in \textbf{bold} and \underline{underline} respectively.}
    \label{tab:main-partial}
    \resizebox{0.8 \textwidth}{!}{%
    \begin{tabular}{@{}l|ccc|ccc|c@{}}
    \toprule
    Dataset          & \multicolumn{3}{c|}{CIFAR-10}        & \multicolumn{3}{c|}{CIFAR-100}       & CUB-200     \\ \midrule
    Partial Ratio $q$   & 0.1        & 0.3        & 0.5        & 0.01       & 0.05       & 0.1        & 0.05       \\ \midrule
    Fully-Supervised & \multicolumn{3}{c|}{94.91\scriptsize{±0.07}}     & \multicolumn{3}{c|}{73.56\scriptsize{±0.10}}      & -           \\ \midrule
    LWS  \citep{wen2021leveraged}            & 90.30\scriptsize{±0.60} & 88.99\scriptsize{±1.43} & 86.16\scriptsize{±0.85} & 65.78\scriptsize{±0.02} & 59.56\scriptsize{±0.33} & 53.53\scriptsize{±0.08} & 39.74\scriptsize{±0.47}                     \\
    PRODEN  \citep{lv2020progressive}         & 90.24\scriptsize{±0.32} & 89.38\scriptsize{±0.31} & 87.78\scriptsize{±0.07} & 62.60\scriptsize{±0.02} & 60.73\scriptsize{±0.03} & 56.80\scriptsize{±0.29} & 62.56\scriptsize{±0.10}                   \\
    CC  \citep{Feng2020ProvablyCP}             & 82.30\scriptsize{±0.21} & 79.08\scriptsize{±0.07} & 74.05\scriptsize{±0.35} & 49.76\scriptsize{±0.45} & 47.62\scriptsize{±0.08} & 35.72\scriptsize{±0.47} & 55.61\scriptsize{±0.02}                      \\
    MSE  \citep{feng2020learning}            & 79.97\scriptsize{±0.45} & 75.65\scriptsize{±0.28} & 67.09\scriptsize{±0.66} & 49.17\scriptsize{±0.05} & 46.02\scriptsize{±1.82} & 43.81\scriptsize{±0.49} & 22.07\scriptsize{±2.36}                     \\
    EXP   \citep{feng2020learning}          & 79.23\scriptsize{±0.10} & 75.79\scriptsize{±0.21} & 70.34\scriptsize{±1.32} & 44.45\scriptsize{±1.50} & 41.05\scriptsize{±1.40} & 29.27\scriptsize{±2.81} & 9.44\scriptsize{±2.32}     \\
    PiCO    \citep{wang2022pico}         & \underline{94.39\scriptsize{±0.18}} & \underline{94.18\scriptsize{±0.12}} & \underline{93.58\scriptsize{±0.06}} & \underline{73.09\scriptsize{±0.34}} & \underline{72.74\scriptsize{±0.30}} & \underline{69.91\scriptsize{±0.24}} & \textbf{72.17\scriptsize{±0.72}}                  \\ 
    \midrule
    Ours & \textbf{96.37\scriptsize{±0.08}} & \textbf{96.26\scriptsize{±0.03}} & \textbf{95.91\scriptsize{±0.05}} & \textbf{75.31\scriptsize{±0.19}} & \textbf{74.58\scriptsize{±0.03}} & \textbf{74.00\scriptsize{±0.02}} &  \underline{70.77\scriptsize{±0.29}}   \\ \bottomrule
    \end{tabular}%
    }
\vspace{-0.1in}
\end{table}

\textbf{Results}. The results for PLL are shown in \cref{tab:main-partial}. 
Our method achieves the best performance compared to the baseline methods. 
Perhaps more surprisingly, on CIFAR-10 and CIFAR-100, our method even outperforms the fully-supervised reference, indicating the potential better generalization capability using the proposed framework, sharing similar insights as in Wu et al. \cite{revisitpllwu22l}.
While PiCO adopts a contrastive learning objective, our method still surpasses PiCO by an average of $\mathbf{2.13\%}$ on CIFAR-10 and $\mathbf{2.72\%}$ on CIFAR-100. Our approach can be further enhanced by incorporating contrastive learning objectives, potentially leading to more significant performance.

\subsection{Semi-Supervised Learning}

\textbf{Setup}. 
For experiments of SSL, we follow the training and evaluation protocols of USB \citep{usb2022} on image and text classification. 
To construct the labeled dataset for semi-supervised learning, we uniformly select $l / C$ samples from each class and treat the remaining samples as the unlabeled dataset. 
We present the results on CIFAR-100 and STL-10 \citep{krizhevsky2009learning} for image classification, and IMDB \citep{maas2011learning} and Amazon Review \citep{mcauley2013hidden} for text classification.
We compare with the current methods with confidence thresholding, such as FixMatch \citep{sohn2020fixmatch}, AdaMatch \citep{berthelot2021adamatch}, FlexMatch \citep{zhang2021flexmatch}, FreeMatch \citep{wang2023freematch}, and SoftMatch \citep{chen2023softmatch}. 
We also compare with methods with the contrastive loss, CoMatch \citep{li2021comatch} and SimMatch \citep{zheng2022simmatch}. 
A full comparison of the USB datasets and hyper-parameters is shown in \cref{sec:append-exp-ssl}.

\begin{table}[t!]
    \centering
    \caption{Error rate of different number of labels $l$ on CIFAR-100, STL-10, IMDB, and Amazon Review datasets for \textbf{semi-supervised learning}.
    }
    \label{tab:main-semi}
    \resizebox{0.95 \textwidth}{!}{%
    \begin{tabular}{@{}lcc|cc|cc|cc@{}}
    \toprule
    \multicolumn{1}{l|}{Datasets} &
      \multicolumn{2}{c|}{CIFAR-100} &
      \multicolumn{2}{c|}{STL-10} &
      \multicolumn{2}{c|}{IMDB} &
      \multicolumn{2}{c}{Amazon Review} \\ \midrule
    \multicolumn{1}{l|}{\# Labels $l$}   & 200         & 400        & 40          & 100         & 20         & 100        & 250        & 1000         \\ \midrule
    \multicolumn{1}{l|}{AdaMatch \citep{berthelot2021adamatch}}    & 22.32\scriptsize{±1.73} & 16.66\scriptsize{±0.62} & 13.64\scriptsize{±2.49}  & \underline{7.62\scriptsize{±1.90}}   & 8.09\scriptsize{±0.99}  & \underline{7.11\scriptsize{±0.20}}    & 45.40\scriptsize{±0.96}   & \textbf{40.16\scriptsize{±0.49}}  \\
    \multicolumn{1}{l|}{FixMatch \citep{sohn2020fixmatch}}    & 29.60\scriptsize{±0.90}   & 19.56\scriptsize{±0.52} & 16.15\scriptsize{±1.89}  & 8.11\scriptsize{±0.68}  & 7.72\scriptsize{±0.33}  & 7.33\scriptsize{±0.13}  & 47.61\scriptsize{±0.83}  & 43.05\scriptsize{±0.54}  \\
    \multicolumn{1}{l|}{FlexMatch \citep{zhang2021flexmatch}}   & 26.76\scriptsize{±1.12} & 18.24\scriptsize{±0.36} & 14.40\scriptsize{±3.11}   & 8.17\scriptsize{±0.78}   & 7.82\scriptsize{±0.77}  & 7.41\scriptsize{±0.38}  & 45.73\scriptsize{±1.60}   & 42.25\scriptsize{±0.33}  \\
    \multicolumn{1}{l|}{CoMatch \citep{li2021comatch}}     & 35.08\scriptsize{±0.69} & 25.35\scriptsize{±0.50}  & 15.12\scriptsize{±1.88}  & 9.56\scriptsize{±1.35}   & \underline{7.44\scriptsize{±0.30}}   & 7.72\scriptsize{±1.14}   &  48.76\scriptsize{±0.90}   & 43.36\scriptsize{±0.21} \\
    \multicolumn{1}{l|}{SimMatch \citep{zheng2022simmatch}}    & 23.78\scriptsize{±1.08} & 17.06\scriptsize{±0.78} & \underline{11.77\scriptsize{±3.20}}  & \textbf{7.55\scriptsize{±1.86}}  & 7.93\scriptsize{±0.55}  & \textbf{7.08\scriptsize{±0.33}}  & 45.91\scriptsize{±0.95}  & 42.21\scriptsize{±0.30}  \\
    \multicolumn{1}{l|}{FreeMatch \citep{wang2023freematch}}   & \textbf{21.40\scriptsize{±0.30}}   & \textbf{15.65\scriptsize{±0.26}} & 12.73\scriptsize{±3.22}  & 8.52\scriptsize{±0.53}   & 8.94\scriptsize{±0.21}  & 7.95\scriptsize{±0.45} & 46.41\scriptsize{±0.60}  & 42.64\scriptsize{±0.06} \\
    \multicolumn{1}{l|}{SoftMatch \citep{chen2023softmatch}}   & 22.67\scriptsize{±1.32} & 16.84\scriptsize{±0.66} & 13.55\scriptsize{±3.16}  & 7.84\scriptsize{±1.72}   & 7.76\scriptsize{±0.58}  & 7.97\scriptsize{±0.72}  & \underline{45.29\scriptsize{±0.95}}  & \underline{42.21\scriptsize{±0.20}} \\ \midrule
    \multicolumn{1}{l|}{Ours} & \underline{22.06\scriptsize{±1.06}} & \underline{16.40\scriptsize{±0.54}} & \textbf{11.09\scriptsize{±0.71}}        & 8.10\scriptsize{±1.02}          & \textbf{7.32\scriptsize{±0.12}} & 7.64\scriptsize{±0.67}  & \textbf{43.96\scriptsize{±0.32}} & 42.32\scriptsize{±0.02}      \\ \bottomrule
    \end{tabular}%
    }
\end{table}

\textbf{Results}. 
We present the results for SSL on \cref{tab:main-semi}. 
Although no individual SSL algorithm dominates the USB benchmark \citep{usb2022}, our method still shows competitive performance. 
Notably, our method performs best on STL-10 with 40 labels and Amazon Review with 250 labels, outperforming the previous best by $\mathbf{0.68\%}$ and $\mathbf{1.33\%}$. 
In the other settings, the performance of our method is also very close to the best-performing methods. 
More remarkably, our method does not employ any thresholding, re-weighting, or contrastive techniques to achieve current results, demonstrating a significant potential to be further explored. 

\subsection{Noisy Label Learning}

\textbf{Setup}. 
We conduct the experiments of NLL following SOP \citep{sopliu22w} on both synthetic symmetric/asymmetric noise on CIFAR-10 and CIFAR-100, and more realistic and larger-scale instance noise on Clothing1M \citep{xiao2015learning}, and WebVision \citep{li2017webvision}. 
To introduce the synthetic symmetric noise to CIFAR-10 and CIFAR-100, we uniformly flip labels for a probability $\eta$ into other classes. 
For asymmetric noise, we only randomly flip the labels for particular pairs of classes. 
The introduced noise is then treated as ground truth labels to train the model.
We mainly select three previous best methods as baselines: DivideMix \citep{Li2020DivideMixLW}; ELR \citep{Liu2020EarlyLearningRP}; and SOP \citep{sopliu22w}.
We also include the normal cross-entropy (CE) training and mixup \citep{zhang2017mixup} as baselines. 
More comparisons of other methods  \citep{Patrini2016MakingDN, Han2018CoteachingRT} and on CIFAR-10N \citep{wei2021learning} with training details and more baselines \citep{jiang2018mentornet,Han2018CoteachingRT} are shown in \cref{sec:append-exp-nll}.

\textbf{Results}. 
We present the noisy label learning results in \cref{tab:main-noise}. 
The proposed method is comparable to the previous best methods. 
On synthetic noise of CIFAR-10, our method demonstrates the best performance on both symmetric noise and asymmetric noise. 
On CIFAR-100, our method generally produces similar results comparable to SOP. 
One may notice that our method shows inferior performance on asymmetric noise of CIFAR-100; we argue this is mainly due to the oversimplification of the noise transition model. 
Our method also achieves the best results on WebVision, outperforming the previous best by $\mathbf{2.05\%}$. 
On Clothing1M, our results are also very close to DivideMix, which trains for 80 epochs compared to 10 epochs in ours.


\begin{table}[t!]
\centering
\caption{Accuracy of synthetic noise on CIFAR-10 and CIFAR-100 and instance noise on Clothing1M and WebVision for \textbf{noisy label learning}. 
We use noise ratio of $\{0.2, 0.5, 0.8\}$ for synthetic symmetric noise and $0.4$ for asymmetric label noise. The instance noise ratio is unknown. 
}
\label{tab:main-noise}
\resizebox{0.98 \textwidth}{!}{%
\begin{tabular}{@{}l|cccc|cccc|c|c@{}}
\toprule
Dataset & \multicolumn{4}{c|}{CIFAR-10}               & \multicolumn{4}{c|}{CIFAR-100}   & Clothing1M & WebVision              \\ \midrule
Noise Type                         & \multicolumn{3}{c}{Sym.} & Asym. & \multicolumn{3}{c}{Sym.} & Asym. & Ins. & Ins. \\ \midrule
Noise Ratio $\eta$                         & 0.2     & 0.5     & 0.8    & 0.4       & 0.2     & 0.5     & 0.8    & 0.4   & - & -   \\ \midrule
CE                       & 87.20    & 80.70    & 65.80   & 82.20      & 58.10    & 47.10    & 23.80   & 43.30   & 69.10 & -    \\
Mixup \citep{zhang2017mixup}                   & 93.50    & 87.90    & 72.30   & -          & 69.90    & 57.30    & 33.60   & -  & - & -       \\
DivideMix \citep{Li2020DivideMixLW}               & 96.10    & 94.60    & 93.20   & 93.40      & 77.10    & 74.60    & 60.20   & 72.10   & \textbf{74.26} & \underline{77.32}   \\
ELR   \citep{Liu2020EarlyLearningRP}                   & 95.80    & 94.80    & 93.30   & 93.00      & \underline{77.70}    & 73.80    & 60.80   & 77.50  & 72.90 & 76.20    \\
SOP \citep{sopliu22w} & \underline{96.30} & \underline{95.50}               & \underline{94.00}               & \underline{93.80}               & \textbf{78.80} & \textbf{75.90} & \underline{63.30}               & \textbf{78.00} & 73.50 & 76.60 \\ \midrule
Ours & \textbf{96.78\scriptsize{±0.11}}    & \textbf{96.60\scriptsize{±0.15}} & \textbf{94.31\scriptsize{±0.07}} & \textbf{94.75\scriptsize{±0.81}} & 77.49\scriptsize{±0.28}     & \underline{75.51\scriptsize{±0.52}}     & \textbf{66.46\scriptsize{±0.72}} & 75.82\scriptsize{±1.89}  & \underline{74.02\scriptsize{±0.12}}  &  \textbf{79.37\scriptsize{±0.09}}  \\ \bottomrule
\end{tabular}%
}
\vspace{-0.1in}
\end{table}

\subsection{Mixed Imprecise Label Learning}
\label{sec:exp-mixed}

\textbf{Setup}. 
We evaluate on CIFAR-10 and CIFAR-100 in a more challenging and realistic setting, the mixture of various imprecise label configurations, with unlabeled, partially labeled, and noisy labeled data existing simultaneously.
We first sample the labeled dataset and treat other samples as the unlabeled.
On the labeled dataset, we generate partial labels and randomly corrupt the true label of the partial labels.  
We set $l \in \{1000, 5000, 50000\}$ for CIFAR-10, and $l \in \{5000, 10000, 50000\}$ for CIFAR-100. 
For partial labels, we set $q \in \{0.1, 0.3, 0.5\}$ for CIFAR-10, and $q \in \{0.01, 0.05, 0.1\}$ for CIFAR-100.
For noisy labels, we set $\eta \in \{0, 0.1, 0.2, 0.3\}$ for both datasets.
Since there is no prior work that can handle all settings all at once, we compare on partial noisy label learning with PiCO+ \citep{wang2022pico+}, IRNet \citep{lian2022irnet}, and DALI \citep{xu2023dali}.
Although there are also prior efforts on partial semi-supervised learning \citep{wang2019partial,wang2020semi}, they do not scale on simple dataset even on CIFAR-10. 
Thus, we did not include them in comparison. 
We conduct additional validation of our method on more complex settings for partial noisy labels with unlabeled data to demonstrate its robustness to various imprecise labels.

\textbf{Results}. 
We report the comparison with partial noisy label learning methods in \cref{tab:main-mix}. 
Compared to previous methods, the proposed method achieves the best performance.
Despite the simplicity, our method outperforms PiCO+ and DALI with mixup, showing the effectiveness of dealing with mixed imprecise labels. 
We also report the results of our methods on more mixed imprecise label configurations in \cref{tab:main-mix-more}. 
Our method demonstrates significant robustness against various settings of the size of labeled data, partial ratio, and noise ratio.
Note that this is the first work that naturally deals with all three imprecise label configurations simultaneously, with superior performance than previous methods handling specific types or combinations of label configurations. This indicates the enormous potential of our work in realistic applications for handling more practical and complicated data annotations common in real world applications.

\begin{table}[t!]
\centering
\caption{Accuracy comparison of \textbf{mixture of different imprecise labels}. We report results of full labels, partial ratio $q$ of 0.1 (0.01) and 0.3 (0.05) for CIFAR-10 (CIFAR-100), and noise ratio $\eta$ of 0.1, 0.2, and 0.3 for CIFAR-10 and CIFAR-100.
}
\label{tab:main-mix}
\resizebox{0.85 \textwidth}{!}{%
\begin{tabular}{@{}l|c|ccc|c|ccc@{}}
\toprule
\multirow{2}{*}{Method} & \multirow{2}{*}{$q$} & \multicolumn{3}{c|}{CIFAR-10, $l$=50000} & \multirow{2}{*}{$q$} & \multicolumn{3}{c}{CIFAR-100, $l$=50000} \\
            &                   & $\eta$=0.1 & $\eta$=0.2 & $\eta$=0.3 &                   & $\eta$=0.1 & $\eta$=0.2 & $\eta$=0.3 \\ \midrule
PiCO+ \citep{wang2022pico+}      & \multirow{6}{*}{0.1} &  93.64   &  93.13   & 92.18 & \multirow{6}{*}{0.01} &   71.42  &   70.22   &  66.14   \\
IRNet \citep{lian2022irnet}      &                   &   93.44  &   92.57  &  92.38   &                   &   71.17  &  70.10   &  68.77   \\
DALI \citep{xu2023dali}       &                   &  94.15   &   94.04  &     93.77   &                   &  72.26   &  71.98   &   71.04   \\
PiCO+ Mixup \citep{xu2023dali} &                   &  94.58   &  94.74   &  94.43 &                   &   75.04  &   74.31  &  71.79   \\
DALI Mixup \citep{xu2023dali}   &                   &  \underline{95.83}    &  \underline{95.86}   &  \underline{95.75}   &                   &  \underline{76.52}   &  \underline{76.55}   &  \underline{76.09}   \\
Ours        &                   &   \textbf{96.47\scriptsize{±0.11}}  &   \textbf{96.09\scriptsize{±0.20}}   &   \textbf{95.83\scriptsize{±0.05}}   &                   &  \textbf{77.53\scriptsize{±0.24}}   &   \textbf{76.96\scriptsize{±0.02}}   &   \textbf{76.43\scriptsize{±0.27}}    \\ \midrule
PiCO+  \citep{wang2022pico+}     & \multirow{6}{*}{0.3} &     92.32  &  92.22   &    89.95  & \multirow{6}{*}{0.05} &    69.40  &  66.67   &     62.24   \\
IRNet  \citep{lian2022irnet}     &                   &   92.81   & 92.18    &  91.35    &                   &    70.73   &  69.33   &   68.09   \\
DALI \citep{xu2023dali}       &                   &  93.44   &  93.25   &    92.42  &                   &  72.28   &   71.35  &   70.05   \\
PiCO+ Mixup \citep{xu2023dali} &                   &  94.02   &  94.03    & 92.94    &                   &  73.06   &   71.37   &  67.56   \\
DALI Mixup  \citep{xu2023dali} &                   &  \underline{95.52}   &  \underline{95.41}    &  \underline{94.67}   &                   &   \underline{76.87}  &  \underline{75.23}   &  \underline{74.49}   \\
Ours        &                   &   \textbf{96.2\scriptsize{±0.02}}   &  \textbf{95.87\scriptsize{±0.14}}   &  \textbf{95.22\scriptsize{±0.06}}   &                   &    \textbf{77.07\scriptsize{±0.16}} &  \textbf{76.34\scriptsize{±0.08}}    & \textbf{75.13\scriptsize{±0.63}}     \\ \bottomrule
\end{tabular}%
}
\vspace{-0.05in}
\end{table}

\begin{table}[t!]
\centering
\caption{Robust test accuracy results of our method on \textbf{more mixture of imprecise label configurations}. 
$l$, $q$ and $\eta$ are the number of labels, partial, and noise ratio. 
}
\label{tab:main-mix-more}
\resizebox{0.9 \textwidth}{!}{%
\begin{tabular}{@{}c|c|cccc|c|c|cccc@{}}
\toprule

\multirow{2}{*}{$l$} & \multirow{2}{*}{$q$} & \multicolumn{4}{c|}{CIFAR10} & \multirow{2}{*}{$l$} & \multirow{2}{*}{$q$} & \multicolumn{4}{c}{CIFAR100} \\
                       &     & $\eta$=0.0 & $\eta$=0.1 & $\eta$=0.2 & $\eta$=0.3 &                         &      & $\eta$=0.0 & $\eta$=0.1 & $\eta$=0.2 & $\eta$=0.3 \\ \midrule
\multirow{3}{*}{5,000} & 0.1 &  95.29\scriptsize{±0.18}   &  93.90\scriptsize{±0.11}   &  92.02\scriptsize{±0.22}   &  89.02\scriptsize{±0.63}   & \multirow{3}{*}{10,000} & 0.01 &  69.90\scriptsize{±0.23}   & 68.74\scriptsize{±0.15}    &  66.87\scriptsize{±0.34}   &  65.34\scriptsize{±0.02}   \\
                       & 0.3 &  95.13\scriptsize{±0.16}   &  92.95\scriptsize{±0.37}    &  90.14\scriptsize{±0.61}   &   87.31\scriptsize{±0.27}    &                         & 0.05 &  69.85\scriptsize{±0.20}    & 68.08\scriptsize{±0.28}    &  66.78\scriptsize{±0.43}   & 64.83\scriptsize{±0.17}     \\
                       & 0.5 &  95.04\scriptsize{±0.10}  &  92.18\scriptsize{±0.52}   & 88.39\scriptsize{±0.62} &  83.09\scriptsize{±0.56}   &                         & 0.10  &  68.92\scriptsize{±0.45}   &  67.15\scriptsize{±0.63}   &   64.44\scriptsize{±1.29}  & 60.26\scriptsize{±1.96}     \\ \midrule
\multirow{3}{*}{1,000} & 0.1 &  94.48\scriptsize{±0.09}   & 91.68\scriptsize{±0.17}    &   87.17\scriptsize{±0.51}  &  81.04\scriptsize{±1.13}   & \multirow{3}{*}{5,000}  & 0.01 & 65.66\scriptsize{±0.27}  &  63.13\scriptsize{±0.27}   &  60.93\scriptsize{±0.17}   &   58.36\scriptsize{±0.56}  \\
                       & 0.3 &    94.35\scriptsize{±0.05}   &  89.94\scriptsize{±1.90}   &  82.06\scriptsize{±1.52}   & 69.20\scriptsize{±2.16}    &                         & 0.05 &  65.06\scriptsize{±0.04}   & 62.28\scriptsize{±0.47}    & 58.92\scriptsize{±0.34} &  53.24\scriptsize{±1.69}    \\
                       & 0.5 &  93.92\scriptsize{±0.29}   &  86.34\scriptsize{±2.37}   &   70.86\scriptsize{±2.78}  &   38.19\scriptsize{±6.55}  &                         & 0.10  &  63.32\scriptsize{±0.55}   &   58.73\scriptsize{±1.33}  &  53.27\scriptsize{±1.57}   & 46.19\scriptsize{±1.04}    \\ \bottomrule
\end{tabular}%
}
\vspace{-0.1in}
\end{table}

\vspace{-.05in}
\section{Conclusion}
\vspace{-0.05in}

We present the imprecise label learning (ILL) framework, a unified and consolidated solution for learning from all types of imprecise labels. 
ILL effectively employs an expectation-maximization (EM) algorithm for maximum likelihood estimation (MLE) of the distribution over the latent ground truth labels $Y$, imprecise label information $I$, and data $X$. 
It naturally extends and encompasses previous formulations for various imprecise label settings, achieving promising results. 
Notably, in scenarios where mixed configurations of imprecise labels coexist, our method exhibits substantial robustness against diverse forms of label imprecision.
The potential \textbf{broader impact} of the ILL framework is substantial. It stands poised to transform domains where obtaining precise labels poses a challenge, offering a simple, unified, and effective approach to such contexts. Beyond the three imprecise label configurations we have demonstrated in this study, the ILL framework shows promise for an extension to more intricate scenarios such as multi-instance learning \citep{ilse2018attention} and multi-label crowd-sourcing learning \citep{ibrahim2023deep}.
However, it is also crucial to acknowledge the \textbf{limitations} of the ILL framework. Although its effectiveness has been substantiated on relatively smaller-scale datasets, additional empirical validation is necessary to assess its scalability to larger datasets.  Furthermore, our study only considers balanced datasets; thus, the performance of the ILL framework when dealing with imbalanced data and open-set data still remains an open area for future exploration.
We hope that our study will constitute a significant stride towards a comprehensive solution for imprecise label learning and catalyze further research in this crucial field.

\section*{Acknowledge}
Masashi Sugiyama was supported by the Institute for AI and Beyond, UTokyo.

\newpage

\bibliography{ref}
\bibliographystyle{unsrt}

\newpage


\appendix

\appendix

\begin{center}
    \Large{\textbf{Appendix}}
\end{center}


\section{Notation}
\label{sec:appen-notation}

We present the notation table for each symbol used in this paper in \cref{tab:append-notation}.

\begin{table}[h!]
\centering
\caption{Notation Table}
\label{tab:append-notation}
\resizebox{0.85 \textwidth}{!}{%
\begin{tabular}{@{}l|l@{}}
\toprule
\multicolumn{1}{c|}{Notation} &\multicolumn{1}{c}{Definition}  \\ \midrule
    $\mathbf{x}$ & A training instance \\
    $y$ & A class index label \\
    $\{\mathbf{x}_i\}_{i \in [N]}$  &  A set of data instances $\mathbf{x}$ of size $N$        \\
    $\{y_i\}_{i \in [N]}$        &   A set of precise label indices $y$ of size $N$          \\ 
    $[\iota]$ & An imprecise label, which might contain multiple class indices \\
    $\{[\iota]_i\}_{i \in [N]}$        &    A set of imprecise labels $[\iota]$ of size $N$     \\ 
    $X$ &   Random variable of training instance       \\
    $\mathcal{X}$  &   Input space where $\mathbf{x}$ is drawn from        \\
    $Y$        &   Random variable of ground-truth labels       \\ 
    $\mathcal{Y}$  &   Label space where $y$ is drawn from        \\
    $I$        &    Random variable of imprecise labels    \\ 
    $f$ & Model backbone \\ 
    $g$ & Model classifier \\
    $h$ & Model multi-layer perceptron \\ 
    $f \circ g$ & Model mapping $\mathcal{X} \rightarrow \mathcal{Y}$ \\ 
    $\theta$ & Learnable parameters of $f \circ g$ \\ 
    $\mathbf{p}(y|\mathbf{x};\theta)$ & Output probability from model $f \circ g$ \\ 
    $f \circ h$ & Model mapping $\mathcal{X} \rightarrow \mathcal{Z}$, where $Z$ is a projected feature space \\ 
    $\mathcal{D}$ & Dataset \\ 
    $\mathcal{L}$ & Loss function \\ 
    $\mathcal{A}_{\mathrm{w}}$ & Weak data augmentation, usually is HorizontalFlip \\
    $\mathcal{A}_{\mathrm{s}}$ & Strong data augmentation, usually is RandAugment \citep{cubuk2020randaugment} \\
    $\mathbf{z}_\mathrm{w}$ & Projected features from $f \circ h$ on weakly-augmented data \\
    $\mathbf{z}_\mathrm{s}$ & Projected features from $f \circ h$ on strongly-augmented data \\
    $\mathcal{M}$ & Memory queue in MoCo \citep{moco2020} \\
    $\mathbf{s}$ &  A partial label, with ground-truth label contained \\
    $\{\mathbf{s}_i\}_{i \in [N]}$ &  A set partial labels, with ground-truth label contained of size $N$ \\
    $S$ & Random variable of partial label \\ 
    $\mathbf{x}^\mathrm{l}$ & A labeled training example \\ 
    $y^\mathrm{l}$ & A labeled class index \\ 
    $\mathbf{x}^\mathrm{u}$ & A unlabeled training example \\ 
    $y^\mathrm{u}$ & A unknown class index for unlabeled data \\ 
    $X^\mathrm{L}$  &   A set of labeled data instances       \\
    $Y^\mathrm{L}$        &   A set of labels for labeled data instances            \\ 
    $X^\mathrm{U}$  &   A set of unlabeled data instances       \\
    $Y^\mathrm{U}$        &   A set of unknown labels for unlabeled data instances            \\ 
    $\hat{p}^\mathbf{u}$ & The maximum predicted probability on unlabeled data $\max(\mathbf{p}(y|\mathbf{x}^\mathrm{u};\theta)$) \\ 
    $\hat{y}^\mathrm{u}$ & The pseudo-label from the predicted probability on unlabeled data $\arg\max(\mathbf{p}(y|\mathbf{x}^\mathrm{u};\theta)$)\\
    $\tau$ & The threshold for confidence thresholding \\ 
    $\hat{y}$ & A corrupted/noisy label \\
    $\hat{y}^{\mathrm{oh}}$ & An one-hot version of the corrupted/noisy label \\
    $\hat{Y}$ & Random variable of noisy labels \\ 
    $\mathbf{u}, \mathbf{v}, \mathbf{m}$  & Noise model related parameters in SOP \citep{sopliu22w} \\
    $\mathcal{T}(\hat{y}|y;\omega)$ & The simplified noise transition model in ILL \\ 
    $\omega$ & The parameters in the simplified noise model \\ 
    \bottomrule
\end{tabular}%
}
\end{table}

\section{Related Work}
\label{sec:related-work}

Many previous methods have been proposed for dealing with the specific types and some combinations of imprecise label configurations.
We revisit the relevant work in this section, especially the state-of-the-art popular baselines for learning with individual and mixture imprecise label configurations.

\textbf{Partial label learning (PLL)}. 
The prior arts can be roughly divided into identification-based for label disambiguation \citep{zhang2015solving,gong2017regularization,xu2019partial,wu2023proper} or average-based for utilizing all candidate labels \citep{hullermeier2006learning,cour2011learning,lv2023robustness}.
The traditional average-based methods usually treat all candidate labels equally, which may involve the misleading false positive labels into training.
To overcome these limitations, researchers have explored identification-based methods, viewing the ground-truth label as a latent variable. They seek to maximize its estimated probability using either the maximum margin criterion \citep{nguyen2008classification,zhang2016partial} or the maximum likelihood criterion \citep{liu2012conditional}.
Deep learning techniques have recently been incorporated into identification-based methods, yielding promising results across multiple datasets. For example, PRODEN \citep{lv2020progressive} proposed a self-training strategy that disambiguates candidate labels using model outputs. CC \citep{Feng2020ProvablyCP} introduced classifier-consistent and risk-consistent algorithms, assuming uniform candidate label generation. LWS \citep{wen2021leveraged} relaxed this assumption and proposed a family of loss functions for label disambiguation. More recently, Wangt et al. \cite{wang2022pico} incorporated contrastive learning into PLL, enabling the model to learn discriminative representations and show promising results under various levels of ambiguity.
RCR involves consistency regularization into PLL recently \citep{revisitpllwu22l}. 

\textbf{Semi-supervised learning (SSL)}. 
SSL is a paradigm for learning with a limited labeled dataset supplemented by a much larger unlabeled dataset.
Consistency regularization and self-training, inspired by clusterness and smoothness assumptions, have been proposed to encourage the network to generate similar predictions for inputs under varying perturbations \citep{tarvainen2017mean,samuli2017temporal,miyato2018virtual}. Self-training \citep{lee2013pseudo,arazo2020pseudo,sohn2020fixmatch} is a widely-used approach for leveraging unlabeled data.
Pseudo Label \citep{lee2013pseudo,arazo2020pseudo}, a well-known self-training technique, iteratively creates pseudo labels that are then used within the same model. 
Recent studies focus largely on generating high-quality pseudo-labels. MixMatch \citep{berthelot2019mixmatch}, for instance, generates pseudo labels by averaging predictions from multiple augmentations. Other methods like ReMixMatch \citep{berthelot2019remixmatch}, UDA \citep{xie2020unsupervised}, and FixMatch \citep{sohn2020fixmatch} adopt confidence thresholds to generate pseudo labels for weakly augmented samples, which are then used to annotate strongly augmented samples.
Methods such as Dash \citep{xu2021dash}, FlexMatch \citep{zhang2021flexmatch}, and FreeMatch \citep{wang2023freematch} dynamically adjust these thresholds following a curriculum learning approach. SoftMatch \citep{chen2023softmatch} introduces a novel utilization of pseudo-labels through Gaussian re-weighting.
SSL has also seen improvements through the incorporation of label propagation, contrastive loss, and meta learning \citep{iscen2019label,pham2021meta,li2021comatch,zheng2022simmatch,usb2022}. 

\textbf{Noisy label learning (NLL)}. 
Overfitting to the noisy labels could result in poor generalization performance, even if the training error is optimized towards zero \citep{zhang2016understanding,zhang2021understanding}. 
Several strategies to address the noisy labels have been proposed \citep{nllsurvey2022}. 
Designing loss functions that are robust to noise is a well-explored strategy for tackling the label noise problem \citep{Zhang2018GeneralizedCE,Wang2019SymmetricCE,ma2020normalized, yu2020learning}. 
Additionally, methods that re-weight loss \citep{liu2016does} have also been explored for learning with noisy labels. 
Another common strategy to handle label noise involves assuming that the noisy label originates from a probability distribution that depends on the actual label.
Early works \citep{Goldberger2016TrainingDN} incorporated these transition probabilities into a noise adaptation layer that is stacked over a classification network and trained in an end-to-end fashion. More recent work, such as Forward \citep{Patrini2016MakingDN}, prefers to estimate these transition probabilities using separate procedures. However, the success of this method is contingent upon the availability of clean validation data \citep{confidentlearn2021} or additional assumptions about the data \citep{zhang2021learning}.
Noise correction has shown promising results in noisy label learning recently \citep{bai2021understanding,li2021learning,li2022selective,sopliu22w}. During the early learning phase, the model can accurately predict a subset of the mislabeled examples \citep{Liu2020EarlyLearningRP}. This observation suggests a potential strategy of correcting the corresponding labels. This could be accomplished by generating new labels equivalent to soft or hard pseudo-labels estimated by the model \citep{tanaka2018joint,yi2019pronoisy}. Co-Teaching uses multiple differently trained networks for correcting noisy labels \citep{Han2018CoteachingRT}. SELFIE \citep{song19b} corrects a subset of labels by replacing them based on past model outputs. Another study in \citep{arazo2019unsupervised} uses a two-component mixture model for sample selection, and then corrects labels using a convex combination.  
Similarly, DivideMix \citep{Li2020DivideMixLW} employs two networks for sample selection using a mixture model and Mixup \citep{zhang2017mixup}.

\textbf{Mixture imprecise label settings}.
Various previous works have explored dealing with distinct types of imprecise labels. However, they have yet to tackle a combination of partial labels, limited labels, and noisy labels, which is a highly realistic scenario. For instance, recent attention has been paid to the issue of partial noisy label learning.
PiCO+ \citep{wang2022pico+}, an extended version of PiCO \citep{wang2022pico}, is tailored specifically for partial noisy labels. 
IRNet \citep{lian2022irnet} uses two modules: noisy sample detection and label correction, transforming the scenario of noisy PLL into a more traditional PLL.
DALI \citep{xu2023dali} is another framework designed to reduce the negative impact of detection errors by creating a balance between the initial candidate set and model outputs, with theoretical assurances of its effectiveness.
Additionally, some work has focused on semi-supervised partial label learning \citep{wang2019partial,wang2020semi}.
No existing research can effectively address the challenge of handling a combination of partial, limited, and noisy labels simultaneously, which underscores the novelty and significance of our work.

\textbf{Previous attempts towards unification of learning from imprecise labels.}
There are earlier attempts for the generalized solutions of different kinds of imprecise labels/observations. 
Den{\oe}ux \cite{denoeux2011maximum} proposed an EM algorithm for the likelihood estimation of fuzzy data and verified the algorithm on linear regression and uni-variate normal mixture estimation. 
Van Rooyen et al. \cite{van2017theory} developed an abstract framework that generically tackles label corruption via the Markov transition.
Quost et al. \cite{quost2016clustering} further extended the EM algorithm of fuzzy data on the finite mixture of Gaussians.
Gong et al. \cite{gong2020centroid} proposed a general framework with centroid estimation for imprecise supervision. 
A unified partial AUC optimization approach was also proposed earlier \citep{xie2024weakly}.
Zhang et al. \cite{zhang2020aggre} and Wei. et al. \cite{uumwei23a} proposed generalized solutions for aggregate observations.
A unified solution based on dynamic programming for count-based weak supervision was also proposed \citep{ShuklaDAE23}
While relating to these works on the surface, ILL does not require any assumption on the imprecise information and generalizes well to more practical settings with noisy labels. 
Some other works for individual settings also related EM framework, but usually involved the approximation on the EM \citep{amini2002semi,alan2016noisyicaspps,wang2022pico}.

\section{Methods}
\label{sec:appendix-method}

\subsection{Derivation of Variational Lower Bound}
\label{sec:append-var-lower-bound}

Evidence lower bound (ELBO), or equivalently variational lower bound \citep{dempster1977maximum}, is the core quantity in EM. 
From \cref{eq:em}, to model $\log P(X, I;\theta)$, we have:
\begin{equation}
\begin{split}
\log P(X, I; \theta) &= \int Q(Y) \log P(X, I;\theta) dY \\
&= \int Q(Y) \log P(X, I;\theta) \frac{P(Y | X, I;\theta)}{P(Y | X, I; \theta)} dY \\
&= \int Q(Y) \log \frac{P(X, I, Y;\theta) Q(Y)}{P(Y | X, I; \theta) Q(Y)} dY  \\
&= \int Q(Y) \log \frac{P(X, I, Y;\theta)}{Q(Y)} dY  - \int Q(Y) \log \frac{P(Y | X, I; \theta)}{Q(Y)} dY \\
\end{split}
\end{equation}
where the first term is the ELBO and the second term is the KL divergence $\mathcal{D}_{KL}(Q(Y) || P(Y | X, I;\theta))$. Replacing $Q(Y)$ with $P(Y|X,I;\theta^t)$ at each iteration will obtain \cref{eq:em}.

\subsection{Instantiations to Partial Label Learning}
\label{sec:append-derive-instan-pll}

The imprecise label $I$ for partial labels is defined as the label candidate sets $S$ with  $\{\mathbf{s}_i\}_{i \in [N]}$ containing the true labels.
Now we can derive \cref{eq:ill-pll} by replacing $I$ with $S$ in \cref{eq:em}:
\begin{equation}
\begin{aligned}
   & \mathbb{E}_{Y|X,I;\theta^t} \left[  \log P(Y|X;\theta) + \log P(I|X, Y; \theta) \right]  \\
   &= \mathbb{E}_{Y|X, S;\theta^t} \left[ \log P(Y | X; \theta) + \log P(I|X, Y; \theta) \right] \\
   &= \sum_{Y} P(Y|X, S; \theta^t) \left[ \log P(Y | X; \theta) + \log P(I|X, Y; \theta) \right]  \\
   &= \sum_{Y} P(Y|X, S; \theta^t) \left[ \log P(Y | X; \theta) \right] + \log P(I | X, Y;\theta)
\end{aligned}
\end{equation}
Note that $P(I|Y, X;\theta)$ can be moved out of the expectation because it is a fixed quantity to any $Y$. 
Now we replace $Y$, $X$, and $S$ to $y$, $\mathbf{x}$, and $\mathbf{s}$ for each instance, and converting the maximization problem to negative log-likelihood minimization problem to drive the loss function:
\begin{equation}
\begin{split}
      \mathcal{L}_{\mathrm{ILL}}^{\mathrm{PLL}}  &= - \frac{1}{N} \sum_i^N  \mathbf{p}(y_i|\mathbf{x}_i, \mathbf{s}_i;\theta^t) \log \mathbf{p}(y_i | \mathbf{x}_i; \theta) - \frac{1}{N} \sum_i^N \log \mathbf{p}( \mathbf{s}_i | \mathbf{x}_i  ,y_i;\theta). \\ 
\end{split}
\end{equation}
The first term is the Cross-Entropy loss we derived in \cref{eq:ill-pll}. 
If $S$ is not instance-dependent, then knowing $Y$ also knows $S$, the second term thus can be ignored in \cref{eq:ill-pll}. 
If $S$ becomes instance-dependent, the second term can be maintained as a supervised term as in \citep{revisitpllwu22l} to optimize $\theta$. 

\subsection{Instantiations to Semi-Supervised Learning}
\label{sec:append-derive-instan-ssl}

In SSL, the input $X$ consists of the labeled data $X^\mathrm{L}$ and the unlabeled data $X^\mathrm{U}$. 
The imprecise label for SSL is realized as the limited number of full labels $Y^\mathrm{L}$ for $X^\mathrm{L}$. 
The labels $Y^\mathrm{U}$ for unlabeled $X^\mathrm{U}$ are unknown and become the latent variable. 
Thus we can write:
\begin{equation}
\begin{aligned}
    & \mathbb{E}_{Y|X,I;\theta^t} \left[  \log P(Y|X;\theta) + \log P(I|X, Y; \theta) \right]  \\ 
    &= \mathbb{E}_{Y^\mathrm{U} |X^\mathrm{U}, X^\mathrm{L},Y^\mathrm{L};\theta^t} \left[  \log P(Y^\mathrm{U}|X^\mathrm{U},X^\mathrm{L};\theta) + \log P(Y^\mathrm{L}|X^\mathrm{L}, X^\mathrm{U}, Y^\mathrm{U}; \theta) \right]  \\
    &= \sum_{Y^\mathrm{U}} P(Y^\mathrm{U} | X^\mathrm{U};\theta^t) \left[  \log P(Y^\mathrm{U}|X^\mathrm{U};\theta) \right] + \log P(Y^\mathrm{L}|X^\mathrm{L}; \theta).   \\
\end{aligned} 
\end{equation}
The negative log-likelihood loss function for $\{\mathbf{x}_i^l, y_i^l\}_{i \in [N^L]}$ and $\{\mathbf{x}^u\}_{i \in [N^U]}$ thus becomes:
\begin{equation}
\begin{split}
    \mathcal{L}_{\mathrm{ILL}}^{\mathrm{SSL}} = \mathcal{L}_{\mathrm{CE}} \left( \mathbf{p}(y|\mathbf{x}^\mathrm{u};\theta)  , \mathbf{p}(y|\mathbf{x}^\mathrm{u};\theta^t)  \right) + \mathcal{L}_{\mathrm{CE}} \left( \mathbf{p}(y|\mathbf{x}^\mathrm{L};\theta), y^\mathrm{L} \right)
\end{split}
\end{equation}

\subsection{Instantiations to Noisy Label Learning}
\label{sec:append-derive-instan-nll}

We denote the given noisy labels as $\hat{Y}$. For noisy label learning, our method naturally supports a noise transition model $\mathcal{T}(\hat{Y} | Y;\omega)$ with learnable parameter $\omega$, as we will show in the following:
\begin{equation}
\begin{split}
     & \mathbb{E}_{Y|X,I;\theta^t} \left[ \log P(Y|X;\theta) + \log P(I|X, Y; \theta) \right]  \\ 
     &= \mathbb{E}_{Y|X, \hat{Y};\theta^t} \left[\log P(Y, \hat{Y} | X;\theta)  \right] \\
     &= \mathbb{E}_{Y|X, \hat{Y};\theta^t} \left[\log P(Y | \hat{Y}, X;\theta) + \log P(\hat{Y} | X;\theta) \right] \\
     &=  \sum_{Y} P(Y | \hat{Y}, X; \theta^t) \log P(Y | \hat{Y}, X;\theta) + \log P(\hat{Y} | X;\theta). \\
\end{split}
\end{equation}
The loss function is:
\begin{equation}
\begin{split}
    \mathcal{L}_{\mathrm{ILL}}^{\mathrm{NLL}} 
    &= \mathcal{L}_{\mathrm{CE}}\left( \mathbf{p}(y|\mathbf{x}, \hat{y};\theta, \omega^t), 
    \mathbf{p}(y| \mathbf{x}, \hat{y}; \theta^t, \omega^t) \right) 
    + \mathcal{L}_{\mathrm{CE}} \left( \mathbf{p}(\hat{y}|\mathbf{x};\theta,\omega), \hat{y} \right)
\end{split}
\end{equation}
Note that both term is computed from the noise transition matrix as mentioned in \cref{eq:ill-nll}.

\subsection{Instantiations to Mixed Imprecise Label Learning}
\label{sec:append-derive-instan}

In this setting, we have both labeled data and unlabeled data, where the labels for the labeled data are both partial and noisy. 
On the unlabeled data, the unsupervised objective is the same as the unsupervised consistency regularization of semi-supervised learning shown in \cref{eq:ill-ssl}. 
On the labeled data, it mainly follows the \cref{eq:ill-nll} of noisy label learning, with the noisy single label becoming the noisy partial labels $\hat{\mathbf{s}}$. 
For noisy partial labels, the noisy supervised objective in Eq. 8 becomes the supervised consistency regularization as in Eq. 6 of partial label setting to train the noise transition model, and the noisy unsupervised objective becomes the consistency regularization of the prediction conditioned on noisy partial labels:
\begin{equation}
\mathcal{L}_{\mathrm{CE}}\left(\mathbf{p}\left(y \mid \mathcal{A}_{\mathrm{s}}(\mathbf{x}), \hat{\mathbf{s}} ; \theta, \omega^t\right), \mathbf{p}\left(y \mid \mathcal{A}_{\mathrm{w}}(\mathbf{x}), \hat{y} ; \theta^t, \omega^t\right)\right)+\mathcal{L}_{\mathrm{CE}}\left(\mathbf{p}\left(\hat{y} \mid \mathcal{A}_{\mathrm{w}}(\mathbf{x}) ; \theta, \omega\right), \hat{\mathbf{s}}\right)
\end{equation}

We can compute both quantity through the noise transition model:
\begin{equation}
    \mathbf{p}(y|\mathbf{x}, \hat{\mathbf{s}};\theta, \omega^t) \propto \mathbf{p}(y|\mathbf{x};\theta) \prod_{\hat{y} \in \hat{\mathbf{s}}} \mathcal{T}(y | \hat{y}; \omega^t) , \text{and }
    \mathbf{p}(\hat{y} |\mathbf{x};\theta,\omega) = \sum_{y \in [C]} \mathbf{p}(y|\mathbf{x};\theta)  \mathcal{T}(\hat{y} | y; \omega).
\end{equation}

\section{Experiments}
\label{sec:appen-exp}

\subsection{Additional Training Details}

We adopt two additional training strategies for the ILL framework. The first is the ``strong-weak'' augmentation strategy \citep{xie2020unsupervised}. Since there is a consistency regularization term in each imprecise label formulation of ILL, we use the soft pseudo-targets of the weakly-augmented data to train the strongly-augmented data. The second is the entropy loss \citep{bridle1991unsup} for class balancing, which is also adopted in SOP \citep{sopliu22w} and FreeMatch \citep{wang2023freematch}. We set the loss weight for the entropy loss uniformly for all experiments as 0.1. 







\subsection{Partial Label Learning}
\label{sec:append-exp-ppl}

\subsubsection{Setup}
Following previous work \citep{xu2022progressive,wen2021leveraged,wang2022pico}, we evaluate our method on partial label learning setting using CIFAR-10, CIFAR-100, and CUB-200 \citep{welinder2010caltech}.
We generate partially labeled datasets by flipping negative labels to false positive labels with a probability $q$, which is also denoted as a partial ratio. 
Specifically, the $C - 1$ negative labels are uniformly aggregated into the ground truth label to form a set of label candidates. 
We consider $q \in \{0.1, 0.3, 0.5\}$ for CIFAR-10, $q \in \{0.01, 0.05, 0.1\}$ for CIFAR-100, and $q=0.05$ for CUB-200. 
For CIFAR-10 and CIFAR-100, we use ResNet-18 \citep{he2016deep} as backbone. We use SGD as an optimizer with a learning rate of $0.01$, a momentum of $0.9$, and a weight decay of $1$e$-3$. 
For CUB-200, we initialize the ResNet-18 \citep{he2016deep} with ImageNet-1K \citep{deng2009imagenet} pre-trained weights. 
We train $800$ epochs for CIFAR-10 and CIFAR-100 \citep{krizhevsky2009learning}, and $300$ epochs for CUB-200, with a cosine learning rate scheduler.
For CIFAR-10 and CIFAR-100, we use an input image size of 32. For CUB-200, we use an input image size of 224. 
A batch size of 256 is used for all datasets. 
The choice of these parameters mainly follows PiCO \citep{wang2022pico}.
We present the full hyper-parameters systematically in \cref{tab:append-param-pll}.

\begin{table}[h!]
\centering
\caption{Hyper-parameters for \textbf{partial label learning} used in experiments.}
\label{tab:append-param-pll}
\resizebox{0.6 \textwidth}{!}{%
\begin{tabular}{@{}c|ccc@{}}
\toprule
\multicolumn{1}{c|}{Hyper-parameter}   & CIFAR-10 & CIFAR-100 & CUB-200 \\ \midrule
Image Size & 32 & 32 & 224 \\
Model      &  ResNet-18 &   ResNet-18        &   \begin{tabular}{c} ResNet-18  \\(ImageNet-1K Pretrained) \end{tabular}     \\
Batch Size        &     256     &    256       &   256      \\
Learning Rate     &     0.01     &   0.01        &   0.01     \\
Weight Decay      &     1e-3     &   1e-3       &   1e-5      \\
LR Scheduler          &     Cosine     &   Cosine   &   Cosine       \\
Training Epochs   &     800    &   800        &  300       \\
Classes           &     10     &     100      &   200      \\ 
\bottomrule
\end{tabular}%
}
\end{table}

\subsubsection{Discussion}
\label{sec:append-exp-rcr}

We additionally compare our method with R-CR \citep{revisitpllwu22l}, which uses a different architecture as the results in \cref{tab:main-partial}. R-CR uses Wide-ResNet34x10 as backbone, and adopts multiple strong data augmentations. It also adjusts the loss weight along training. 
For fair comparison, we use the same architecture without multiple augmentation and the curriculum adjust on loss. 
The results are shown in \cref{tab:pll-rcr}, where our method outperforms R-CR on CIFAR-10 and is comparable on CIFAR-100.

\begin{table}[h]
\centering
\caption{Comparison with R-CR in partial label learning}
\label{tab:pll-rcr}
\resizebox{0.5 \textwidth}{!}{%
\begin{tabular}{@{}c|cc|cc@{}}
\toprule
\multirow{2}{*}{Method} &
  \multicolumn{2}{c|}{CIFAR-10} &
  \multicolumn{2}{c}{CIFAR-100} \\
 &
  0.3 &
  0.5 &
  0.05 &
  0.10 \\ \midrule
R-CR &
  97.28\scriptsize{$\pm$0.02} &
  97.05\scriptsize{$\pm$0.05} &
  82.77\scriptsize{$\pm$0.10} &
  82.24\scriptsize{$\pm$0.07} \\
Ours &
  97.55\scriptsize{$\pm$0.07} &
  97.17\scriptsize{$\pm$0.11} &
  82.46\scriptsize{$\pm$0.08} &
  82.22\scriptsize{$\pm$0.05} \\ \bottomrule
\end{tabular}%
}
\end{table}

We also provide the comparison of our method on instance-dependent partial label learning as proposed by Xu et al. \cite{xu2021instance,xu2022progressive}. 
Due to the nature of instance-dependence, we maintain the term $P(S|Y,X;\theta)$ from \cref{eq:em} as a supervised term for optimization. 
We compare our method with VALEN \cite{xu2021instance}, RCR \cite{revisitpllwu22l}, PiCO \cite{wang2022pico}, and POP \cite{xu2022progressive} on MNIST, Kuzushiji-MNIST, Fashion-MNIST, CIFAR-10, and CIFAR-100, with synthetic instance-dependent partial labels generated according to Xu et al. \cite{xu2022progressive}. 
From the results in \cref{tab:ins-pll}, we proposed method demonstrate the best performance across different datasets evaluated.

\begin{table}[h]
\centering
\caption{Comparison on instance-dependent partial label learning}
\resizebox{0.8 \textwidth}{!}{%
\begin{tabular}{@{}c|ccccc@{}}
\toprule
      & MNIST & Kuzushiji-MNIST & Fashion-MNIST & CIFAR-10 & CIFAR-100 \\ \midrule
VALEN \citep{xu2021instance} & 99.03 & 90.15           & 96.31         & 92.01    & 71.48     \\
RCR \citep{revisitpllwu22l}   & 98.81 & 90.62           & 96.64         & 86.11    & 71.07     \\
PiCO \citep{wang2022pico} & 98.76 & 88.87           & 94.83         & 89.35    & 66.30     \\
POP \citep{xu2022progressive}   & \textbf{99.28} & 91.09           & 96.93         & 93.00    & 71.82     \\ \midrule
Ours  & 99.19 & \textbf{91.35}           & \textbf{97.01}         & \textbf{93.86}    & \textbf{72.43}     \\ \bottomrule
\end{tabular}%
}
\label{tab:ins-pll}
\end{table}

A recent work on PLL discussed and analyzed the robustness performance of different loss functions, especially the average-based methods \citep{lv2023robustness}. 
We perform a similar analysis here for the derived loss function in ILL. 
Following the notation in \cite{lv2023robustness}, let $\mathbf{s}$ denote the candidate label set, $\mathbf{x}$ as the training instance, $g$ as the probability score from the model, and $f$ as the classifier $f(\boldsymbol{x})=\underset{i \in \mathcal{Y}}{\arg \max } g_i(\boldsymbol{x})$, the average-based PLL can be formulated as:
\begin{equation}
    \mathcal{L}_{avg-PLL}(f(\boldsymbol{x}), \mathbf{s}) = \frac{1}{|\mathbf{s}|} \sum_{i \in \mathbf{s}} \ell(f(\boldsymbol{x}), i)
\end{equation}
Lv et al. \cite{lv2023robustness} compared different loss functions $\ell$ on both noise-free and noisy PLL settings, where they find both theoretically and empirically that average-based PLL with \textit{bounded} loss are robust under mild assumptions. Empirical study in \cite{lv2023robustness} suggests that both \textit{Mean Absolute Error} and \textit{Generalized Cross-Entropy} loss \citep{Zhang2018GeneralizedCE} that proposed for noisy label learning achieves the best performance and robustness for average-based PLL. 

Our solution for PLL can be viewed as an instantiation of the average-based PLL as in \citep{lv2023robustness} with:
\begin{equation}
\ell(f(\mathbf{x}), i) = - \bar{g}_i(\mathbf{x}) \log g_i(\mathbf{x})
\end{equation}

where $\bar{g}$ is normalized probability over $\mathbf{s}$ with detached gradient. We can further show that the above loss function is bounded for $0 < \ell \leq \frac{1}{e}$ and thus bounded for summation of all classes, which demonstrates robustness, as we show in \cref{tab:main-mix}.




\subsection{Semi-Supervised Learning}
\label{sec:append-exp-ssl}

\subsubsection{Setup}

For experiments of SSL, we follow the training and evaluation protocols of USB \citep{usb2022} on image and text classification. 
To construct the labeled dataset for semi-supervised learning, we uniformly select $l / C$ samples from each class and treat the remaining samples as the unlabeled dataset. 
For image classification tasks, ImageNet-1K \citep{deng2009imagenet} Vision Transformers \citep{dosovitskiy2020image} are used, including CIFAR-100 \citep{krizhevsky2009learning}, EuroSAT \citep{helber2019eurosat}, STL-10 \citep{coates2011analysis}, TissueMNIST \citep{medmnistv1,medmnistv2}, Semi-Aves \citep{su2021semi}.
For text classification tasks, we adopt BERT \citep{devlin2018bert} as backbone, including IMDB \citep{maas2011learning}, Amazon Review \citep{mcauley2013hidden}, Yelp Review \citep{yelpwebsite}, AG News \citep{zhang2015character} , Yahoo Answer \citep{chang2008importance}. 
The hyper-parameters strictly follow USB, and are shown in \cref{tab:append-hyper-cv} and \cref{tab:append-hyper-nlp}.

\begin{table}[!htbp]
\centering
\caption{Hyper-parameters of \textbf{semi-supervised learning} used in vision experiments of USB.}
\resizebox{0.85\textwidth}{!}{
\begin{tabular}{c|ccccc} \toprule
Hyper-parameter & CIFAR-100 & STL-10 & Euro-SAT & TissueMNIST & Semi-Aves \\ 
\midrule
Image Size  & 32 & 96 & 32 & 32 & 224 \\ 
Model    &  ViT-S-P4-32 & ViT-B-P16-96 & ViT-S-P4-32 & ViT-T-P4-32 & ViT-S-P16-224  \\ 
Labeled Batch size & \multicolumn{5}{c}{16} \\ 
Unlabeled Batch size & \multicolumn{5}{c}{16} \\ 
Learning Rate & 5e-4 & 1e-4 & 5e-5  & 5e-5  & 1e-3 \\ 
Weight Decay &  \multicolumn{5}{c}{5e-4} \\ 
Layer Decay Rate & 0.5 & 0.95 &  1.0 &  0.95 & 0.65  \\ 
LR Scheduler & \multicolumn{5}{c}{$\eta = \eta_0 \cos(\frac{7\pi k}{16K})$} \\ 
Training epochs & \multicolumn{5}{c}{20} \\ 
Classes & 100 & 10 & 10 & 10 & 200 \\ 
Model EMA Momentum & \multicolumn{5}{c}{0.0}\\ 
Prediction EMA Momentum & \multicolumn{5}{c}{0.999}\\ 
Weak Augmentation & \multicolumn{5}{c}{Random Crop, Random Horizontal Flip} \\ 
Strong Augmentation & \multicolumn{5}{c}{RandAugment \citep{cubuk2020randaugment}} \\
\bottomrule
\end{tabular}
}
\label{tab:append-hyper-cv}
\end{table}

\begin{table}[!htbp]
\centering
\caption{Hyper-parameters of \textbf{semi-supervised learning} NLP experiments in USB.}
\resizebox{0.75\textwidth}{!}{
\begin{tabular}{c|ccccc}\toprule
Hyper-parameter &  AG News & Yahoo! Answer & IMDB & Amazom-5 & Yelp-5 \\ 
\midrule
Max Length    &  \multicolumn{5}{c}{512} \\ 
Model    &  \multicolumn{5}{c}{Bert-Base} \\ 
Labeled Batch size & \multicolumn{5}{c}{4} \\ 
Unlabeled Batch size & \multicolumn{5}{c}{4} \\ 
Learning Rate & 5e-5 & 1e-4 & 5e-5 & 1e-5 & 5e-5 \\ 
Weight Decay&  \multicolumn{5}{c}{1e-4} \\ 
Layer Decay Rate & 0.65 & 0.65 & 0.75 & 0.75 & 0.75 \\ 
LR Scheduler & \multicolumn{5}{c}{$\eta = \eta_0 \cos(\frac{7\pi k}{16K})$} \\ 
Training epochs & \multicolumn{5}{c}{10} \\ 
Classes & 4 & 10 & 2 & 5 & 5 \\ 
Model EMA Momentum & \multicolumn{5}{c}{0.0}\\ 
Prediction EMA Momentum & \multicolumn{5}{c}{0.999}\\ 
Weak Augmentation & \multicolumn{5}{c}{None} \\ 
Strong Augmentation & \multicolumn{5}{c}{Back-Translation \citep{xie2020unsupervised}} \\
\bottomrule
\end{tabular}
}
\label{tab:append-hyper-nlp}
\end{table}



\begin{table}[h!]
\centering
\caption{Error rate comparison of different number of labels on CIFAR-100, STL-10, EuroSAT, TissueMNIST, and SemiAves for \textbf{semi-supervised learning}. We use USB \citep{usb2022} image classification task results. The best results are indicated in bold. Our results are averaged over 3 independent runs.}
\label{tab:append-ssl-cv}
\resizebox{\textwidth}{!}{%
\begin{tabular}{@{}lcc|cc|cc|cc|c@{}}
\toprule
\multicolumn{1}{l|}{Datasets} &
  \multicolumn{2}{c|}{CIFAR-100} &
  \multicolumn{2}{c|}{STL-10} &
  \multicolumn{2}{c|}{EuroSat} &
  \multicolumn{2}{c|}{TissueMNIST} &
  SemiAves \\ \midrule
\multicolumn{1}{l|}{\# Labels}   & 200        & 400        & 40          & 100         & 20         & 40         & 80         & 400        & 3959       \\ \midrule
\multicolumn{1}{l|}{Pseudo-Label \citep{lee2013pseudo}} & 33.99\scriptsize{±0.95} & 25.32\scriptsize{±0.29} & 19.14\scriptsize{±1.33}   & 10.77\scriptsize{±0.60}   & 25.46\scriptsize{±1.36} & 15.70\scriptsize{±2.12}  & 56.92\scriptsize{±4.54} & 50.86\scriptsize{±1.79} & 40.35\scriptsize{±0.30}  \\
\multicolumn{1}{l|}{Mean-Teacher \citep{tarvainen2017mean}} & 35.47\scriptsize{±0.40}  & 26.03\scriptsize{±0.30}  & 18.67\scriptsize{±1.69}  & 24.19\scriptsize{±10.15} & 26.83\scriptsize{±1.46} & 15.85\scriptsize{±1.66} & 62.06\scriptsize{±3.43} & 55.12\scriptsize{±2.53} & 38.55\scriptsize{±0.21} \\
\multicolumn{1}{l|}{VAT \citep{miyato2018virtual}}         & 31.49\scriptsize{±1.33}  & 21.34\scriptsize{±0.50}   & 18.45\scriptsize{±1.47}   & 10.69\scriptsize{±0.51}   & 26.16\scriptsize{±0.96}  & 10.09\scriptsize{±0.94} & 57.49\scriptsize{±5.47}  & 51.30\scriptsize{±1.73}   & 38.82\scriptsize{±0.04}  \\
\multicolumn{1}{l|}{MixMatch \citep{berthelot2019mixmatch}}    & 38.22\scriptsize{±0.71} & 26.72\scriptsize{±0.72} & 58.77\scriptsize{±1.98}  & 36.74\scriptsize{±1.24}  & 24.85\scriptsize{±4.85} & 17.28\scriptsize{±2.67} & 55.53\scriptsize{±1.51} & 49.64\scriptsize{±2.28} & 37.25\scriptsize{±0.08} \\
\multicolumn{1}{l|}{ReMixMatch \citep{berthelot2019remixmatch}}  & 22.21\scriptsize{±2.21}  & 16.86\scriptsize{±0.57}  & 13.08\scriptsize{±3.34}   & \textbf{7.21\scriptsize{±0.39}}    & \textbf{5.05\scriptsize{±1.05}}   & 5.07\scriptsize{±0.56}   & 58.77\scriptsize{±4.43}  & 49.82\scriptsize{±1.18}  & \textbf{30.20\scriptsize{±0.03}}   \\
\multicolumn{1}{l|}{AdaMatch \citep{berthelot2021adamatch}}    & 22.32\scriptsize{±1.73} & 16.66\scriptsize{±0.62} & 13.64\scriptsize{±2.49}  & 7.62\scriptsize{±1.90}    & 7.02\scriptsize{±0.79}  & \textbf{4.75\scriptsize{±1.10}}   & 58.35\scriptsize{±4.87} & 52.40\scriptsize{±2.08}  & 31.75\scriptsize{±0.13} \\
\multicolumn{1}{l|}{FixMatch \citep{sohn2020fixmatch}}    & 29.60\scriptsize{±0.90}   & 19.56\scriptsize{±0.52} & 16.15\scriptsize{±1.89}  & 8.11\scriptsize{±0.68}  & 13.44\scriptsize{±3.53} & 5.91\scriptsize{±2.02}  & \textbf{55.37\scriptsize{±4.50}}  & 51.24\scriptsize{±1.56} & 31.90\scriptsize{±0.06}  \\
\multicolumn{1}{l|}{FlexMatch \citep{zhang2021flexmatch}}   & 26.76\scriptsize{±1.12} & 18.24\scriptsize{±0.36} & 14.40\scriptsize{±3.11}   & 8.17\scriptsize{±0.78}   & 5.17\scriptsize{±0.57}  & 5.58\scriptsize{±0.81}  & 58.36\scriptsize{±3.80}  & 51.89\scriptsize{±3.21} & 32.48\scriptsize{±0.15} \\
\multicolumn{1}{l|}{Dash \citep{xu2021dash}}        & 30.61\scriptsize{±0.98} & 19.38\scriptsize{±0.10}  & 16.22\scriptsize{±5.95}  & 7.85\scriptsize{±0.74}   & 11.19\scriptsize{±0.90}  & 6.96\scriptsize{±0.87}  & 56.98\scriptsize{±2.93} & 51.97\scriptsize{±1.55} & 32.38\scriptsize{±0.16} \\
\multicolumn{1}{l|}{CoMatch \citep{li2021comatch}}     & 35.08\scriptsize{±0.69} & 25.35\scriptsize{±0.50}  & 15.12\scriptsize{±1.88}  & 9.56\scriptsize{±1.35}   & 5.75\scriptsize{±0.43}  & 4.81\scriptsize{±1.05}  & 59.04\scriptsize{±4.90}  & 52.92\scriptsize{±1.04} & 38.65\scriptsize{±0.18} \\
\multicolumn{1}{l|}{SimMatch \citep{zheng2022simmatch}}    & 23.78\scriptsize{±1.08} & 17.06\scriptsize{±0.78} & 11.77\scriptsize{±3.20}  & 7.55\scriptsize{±1.86}   & 7.66\scriptsize{±0.60}  & 5.27\scriptsize{±0.89}  & 60.88\scriptsize{±4.31} & 52.93\scriptsize{±1.56} & 33.85\scriptsize{±0.08} \\
\multicolumn{1}{l|}{FreeMatch \citep{wang2023freematch}}   & \textbf{21.40\scriptsize{±0.30}}   & \textbf{15.65\scriptsize{±0.26}} & 12.73\scriptsize{±3.22}  & 8.52\scriptsize{±0.53}   & 6.50\scriptsize{±0.78}   & 5.78\scriptsize{±0.51}  & 58.24\scriptsize{±3.08} & 52.19\scriptsize{±1.35} & 32.85\scriptsize{±0.31} \\
\multicolumn{1}{l|}{SoftMatch \citep{chen2023softmatch}}   & 22.67\scriptsize{±1.32} & 16.84\scriptsize{±0.66} & 13.55\scriptsize{±3.16}  & 7.84\scriptsize{±1.72}   & 5.75\scriptsize{±0.62}  & 5.90\scriptsize{±1.42}   & 57.98\scriptsize{±3.66} & 51.73\scriptsize{±2.84} & 31.80\scriptsize{±0.22}  \\ \midrule
\multicolumn{1}{l|}{Ours} & 22.06\scriptsize{±1.06} & 17.40\scriptsize{±1.04} & \textbf{11.09\scriptsize{±0.71}}        & 8.10\scriptsize{±1.02}          & 5.86\scriptsize{±1.06}  & 5.74\scriptsize{±1.13}  & 57.99\scriptsize{±2.16}       & \textbf{50.95\scriptsize{±2.03}}       & 33.08\scriptsize{±0.26} \\ \bottomrule
\end{tabular}%
}
\end{table}

\begin{table}[h!]
\centering
\caption{Error rate comparison of different number of labels on IMDB, AG News, Amazon Review, Yahoo Answers, and Yelp Review for \textbf{semi-supervised learning}. We use USB \citep{usb2022} text classification task results. Best results are indicated in bold. Our results are averaged over 3 independent runs.}
\label{tab:append-ssl-nlp}
\resizebox{\textwidth}{!}{%
\begin{tabular}{@{}l|cc|cc|cc|cc|cc@{}}
\toprule
Datasets &
  \multicolumn{2}{c|}{IMDB} &
  \multicolumn{2}{c|}{AG News} &
  \multicolumn{2}{c|}{Amazon Review} &
  \multicolumn{2}{c|}{Yahoo Answers} &
  \multicolumn{2}{c}{Yelp Review} \\ \midrule
\# Labels   & 20         & 100        & 40          & 200         & 250         & 1000        & 500        & 2000       & 250         & 1000        \\ \midrule
Pseudo-Label \citep{lee2013pseudo} & 45.45\scriptsize{±4.43} & 19.67\scriptsize{±1.01} & 19.49\scriptsize{±3.07}  & 14.69\scriptsize{±1.88}  & 53.45\scriptsize{±1.9}   & 47.00\scriptsize{±0.79}   & 37.70\scriptsize{±0.65}  & 32.72\scriptsize{±0.31} & 54.51\scriptsize{±0.82} & 47.33\scriptsize{±0.20}   \\
Mean-Teacher \citep{tarvainen2017mean} & 20.06\scriptsize{±2.51} & 13.97\scriptsize{±1.49} & 15.17\scriptsize{±1.21} & 13.93\scriptsize{±0.65} & 52.14\scriptsize{±0.52} & 47.66\scriptsize{±0.84} & 37.09\scriptsize{±0.18} & 33.43\scriptsize{±0.28} & 50.60\scriptsize{±0.62} &  47.21\scriptsize{±0.31} \\
VAT \citep{miyato2018virtual}         & 25.93\scriptsize{±2.58} & 11.61\scriptsize{±1.79} & 14.70\scriptsize{±1.19}   & 11.71\scriptsize{±0.84}  & 49.83\scriptsize{±0.46}  & 46.54\scriptsize{±0.31}  & 34.87\scriptsize{±0.41} & 31.50\scriptsize{±0.35}  & 52.97\scriptsize{±1.41}  & 45.30\scriptsize{±0.32}  \\
MixMatch \citep{berthelot2019mixmatch}    & 26.12\scriptsize{±6.13} & 15.47\scriptsize{±0.65} & 13.50\scriptsize{±1.51}   & 11.75\scriptsize{±0.60}   & 59.54\scriptsize{±0.67} & 61.69\scriptsize{±3.32}  & 35.75\scriptsize{±0.71} & 33.62\scriptsize{±0.14} & 53.98\scriptsize{±0.59}  & 51.70\scriptsize{±0.68}  \\
AdaMatch \citep{berthelot2021adamatch}    & 8.09\scriptsize{±0.99}  & 7.11\scriptsize{±0.20}  & \textbf{11.73\scriptsize{±0.17}}  & \textbf{11.22\scriptsize{±0.95}}  & 46.72\scriptsize{±0.72}  & 42.27\scriptsize{±0.25}  & 32.75\scriptsize{±0.35} & 30.44\scriptsize{±0.31} & 45.40\scriptsize{±0.96}   & 40.16\scriptsize{±0.49}  \\
FixMatch \citep{sohn2020fixmatch}    & 7.72\scriptsize{±0.33}  & 7.33\scriptsize{±0.13}  & 30.17\scriptsize{±1.87}  & 11.71\scriptsize{±1.95} & 47.61\scriptsize{±0.83}  & 43.05\scriptsize{±0.54}  & \textbf{33.03\scriptsize{±0.49}} & 30.51\scriptsize{±0.53} & 46.52\scriptsize{±0.94}  & 40.65\scriptsize{±0.46}  \\
FlexMatch \citep{zhang2021flexmatch}   & 7.82\scriptsize{±0.77}  & 7.41\scriptsize{±0.38}  & 16.38\scriptsize{±3.94}  & 12.08\scriptsize{±0.73}  & 45.73\scriptsize{±1.60}   & 42.25\scriptsize{±0.33}  & 35.61\scriptsize{±1.08} & 31.13\scriptsize{±0.18} & \textbf{43.35\scriptsize{±0.69}}  & 40.51\scriptsize{±0.34} \\
Dash \citep{xu2021dash}        & 8.34\scriptsize{±0.86}  & 7.55\scriptsize{±0.35}  & 17.67\scriptsize{±3.19} & 13.76\scriptsize{±1.67}  & 47.10\scriptsize{±0.74 }  & 43.09\scriptsize{±0.60}   & 35.26\scriptsize{±0.33} & 31.19\scriptsize{±0.29} & 45.24\scriptsize{±2.02}  & 40.14\scriptsize{±0.79}  \\
CoMatch \citep{li2021comatch}     & 7.44\scriptsize{±0.30}  & 7.72\scriptsize{±1.14}  & 11.95\scriptsize{±0.76}  & 10.75\scriptsize{±0.35}  & 48.76\scriptsize{±0.90}   & 43.36\scriptsize{±0.21}  & 33.48\scriptsize{±0.51} & 30.25\scriptsize{±0.35} & 45.40\scriptsize{±1.12}   & 40.27\scriptsize{±0.51}  \\
SimMatch \citep{zheng2022simmatch}    & 7.93\scriptsize{±0.55}  & \textbf{7.08\scriptsize{±0.33}}  & 14.26\scriptsize{±1.51}  & 12.45\scriptsize{±1.37}  & 45.91\scriptsize{±0.95}  & 42.21\scriptsize{±0.30}   & 33.06\scriptsize{±0.20}  & 30.16\scriptsize{±0.21} & 46.12\scriptsize{±0.48}  & 40.26\scriptsize{±0.62}  \\
FreeMatch \citep{wang2023freematch}   & 8.94\scriptsize{±0.21}  & 7.95\scriptsize{±0.45} & 12.98\scriptsize{±0.58}  & 11.73\scriptsize{±0.63}  & 46.41\scriptsize{±0.60}   & 42.64\scriptsize{±0.06}  & 32.77\scriptsize{±0.26} & 30.32\scriptsize{±0.18} & 47.95\scriptsize{±1.45}  & 40.37\scriptsize{±1.00}   \\
SoftMatch \citep{chen2023softmatch}   & 7.76\scriptsize{±0.58}  & 7.97\scriptsize{±0.72}  & 11.90\scriptsize{±0.27}   & 11.72\scriptsize{±1.58}  & 45.29\scriptsize{±0.95}  & \textbf{42.21\scriptsize{±0.20}}   & 33.07\scriptsize{±0.31} & 30.44\scriptsize{±0.62} & 44.09\scriptsize{±0.50}   & 39.76\scriptsize{±0.13}  \\ \midrule
Ours  & \textbf{7.32\scriptsize{±0.12}} & 7.64\scriptsize{±0.67} & 14.77\scriptsize{±1.59} & 12.21\scriptsize{±0.82} & \textbf{43.96\scriptsize{±0.32}} & 42.32\scriptsize{±0.02} & 33.80\scriptsize{±0.25} & 30.86\scriptsize{±0.17} & 44.82\scriptsize{±0.17} & \textbf{39.67\scriptsize{±0.71}} \\ \bottomrule
\end{tabular}%
}
\end{table}

\subsubsection{Results}

In the main paper, we only provide the comparison on CIFAR-100, STL-10, IMDB, and Amazon Review. 
Here we provide the full comparison in \cref{tab:append-ssl-cv} and \cref{tab:append-ssl-nlp}. 
From the full results, similar conclusion can be drawn as in the main paper. 
Our ILL framework demonstrates comparable performance as previous methods. 

\subsection{Noisy Label Learning}
\label{sec:append-exp-nll}

\subsubsection{Setup}

We conduct experiments of noisy label learning following SOP \citep{sopliu22w}. 
We evaluate the proposed method on both synthetic symmetric/asymmetric noise on CIFAR-10 and CIFAR-100, and more realistic and larger-scale instance noise on Clothing1M and WebVision. 
To introduce the synthetic symmetric noise to CIFAR-10 and CIFAR-100, we uniformly flip labels for a probability $\eta$ into other classes. 
For asymmetric noise, we only randomly flip the labels for particular pairs of classes. 
For CIFAR-10 and CIFAR-100, we train PreAct-ResNet-18 with SGD using a learning rate of $0.02$, a weight decay of $1e-3$, and a momentum of $0.9$. We train for $300$ epochs with a cosine learning rate schedule and a batch size of 128. 
For WebVision, we use InceptionResNet-v2 as the backbone and set the batch size to $32$. Other settings are similar to CIFAR-10. 
For Clothing1M, we use ImageNet-1K pre trained ResNet-50 as the backbone. We train it using SGD with an initial learning rate of $2e$-$3$ for a total of 10 epochs, where the learning rate is reduced by 10 after 5 epochs. 
In addition, we also conduct experiments on CIFAR-10N and CIFAR-100N. We present the detailed hyper-parameters in \cref{tab:append-param-nll}.

\begin{table}[h!]
\centering
\caption{Hyper-parameters for \textbf{noisy label learning} used in experiments.}
\label{tab:append-param-nll}
\resizebox{0.9 \textwidth}{!}{%
\begin{tabular}{@{}c|cccc@{}}
\toprule
\multicolumn{1}{c|}{Hyper-parameter}   & CIFAR-10 (CIFAR-10N) & CIFAR-100 (CIFAR-100N) & Clothing1M & WebVision \\ \midrule
Image Size & 32 & 32 & 224 & 299 \\
Model      &  PreAct-ResNet-18 (ResNet-34) &   PreAct-ResNet-18 (ResNet-34)     &  \begin{tabular}{c} ResNet-50  \\(ImageNet-1K Pretrained) \end{tabular}  & Inception-ResNet-v2   \\
Batch Size        &     128     &    128       &   64  & 32    \\
Learning Rate     &     0.02     &   0.02        &   0.002 &   0.02 \\
Weight Decay      &     1e-3     &   1e-3       &   1e-3   & 5e-4  \\
LR Scheduler          &     Cosine     &   Cosine   &   MultiStep     &  MultiStep \\
Training Epochs   &     300    &   300        &  10    & 100  \\
Classes           &     10     &     100      &   14    & 50  \\ 
Noisy Matrix Scale & 1.0 & 2.0 & 0.5 & 2.5\\
\bottomrule
\end{tabular}%
}
\end{table}



\subsubsection{Results}

In addition to the results regarding noisy label learning provided in the main paper, we also present comparison results on CIFAR-10N and CIFAR-100N \citep{wei2021learning} in \cref{tab:append-noisy-n}. We include a full comparison on Clothing1M and WebVision, incorporating methods like Co-Teaching, Forward, and CORES, in \cref{tab:append-noisy-web}. 
As shown in \cref{tab:append-noisy-n}, the proposed ILL framework achieves performance comparable to the previous best method, SOP \citep{sopliu22w}.
On CIFAR-10N, our method yields results very close to SOP in the Random and Aggregate case noise scenarios and surpasses SOP in the Worst case noise scenario. 
However, on CIFAR-100N, our method slightly underperforms previous methods, possibly due to the oversimplified noise model utilized in ILL. 
We believe that a more realistic noise transition model and further tuning of our method could lead to improved performance.

\begin{table}[h!]
\centering
\caption{Test accuracy comparison of instance independent label noise on CIFAR-10N and CIFAR-100N for \textbf{noisy label learning}. 
The best results are indicated in \textbf{bold}, and the second best results are indicated in \underline{underline}. Our results are averaged over three independent runs with  ResNet34 as the backbone.}
\label{tab:append-noisy-n}
\resizebox{0.96 \textwidth}{!}{%
\begin{tabular}{@{}l|cccccc|cc@{}}
\toprule
Dataset & \multicolumn{6}{c|}{CIFAR-10N}                                                                                 & \multicolumn{2}{c}{CIFAR-100N} \\ \midrule
Noisy Type & Clean      & Random 1   & Random 2   & Random 3   & Aggregate  & Worst      & Clean               & Noisy               \\ \midrule
CE                    & 92.92\scriptsize{±0.11} & 85.02\scriptsize{±0.65} & 86.46\scriptsize{±1.79} & 85.16\scriptsize{±0.61} & 87.77\scriptsize{±0.38}                         & 77.69\scriptsize{±1.55}            & 76.70\scriptsize{±0.74}     & 55.50\scriptsize{±0.66}    \\
Forward \citep{Patrini2016MakingDN}              & 93.02\scriptsize{±0.12} & 86.88\scriptsize{±0.50} & 86.14\scriptsize{±0.24} & 87.04\scriptsize{±0.35} & 88.24\scriptsize{±0.22}                         & 79.79\scriptsize{±0.46}            & 76.18\scriptsize{±0.37}     & 57.01\scriptsize{±1.03}    \\
Co-teaching \citep{Han2018CoteachingRT}           & 93.35\scriptsize{±0.14} & 90.33\scriptsize{±0.13} & 90.30\scriptsize{±0.17} & 90.15\scriptsize{±0.18} & 91.20\scriptsize{±0.13}     & 83.83\scriptsize{±0.13}            & 73.46\scriptsize{±0.09}     & 60.37\scriptsize{±0.27}    \\
DivideMix  \citep{Li2020DivideMixLW}           & -          & \underline{95.16\scriptsize{±0.19}} & \underline{95.23\scriptsize{±0.07}} & \underline{95.21\scriptsize{±0.14}} &  95.01\scriptsize{±0.71} & 92.56\scriptsize{±0.42}            & -              & 71.13\scriptsize{±0.48}    \\
ELR \citep{Liu2020EarlyLearningRP}                  & 95.39\scriptsize{±0.05} & 94.43\scriptsize{±0.41} & 94.20\scriptsize{±0.24} & 94.34\scriptsize{±0.22} & 94.83\scriptsize{±0.10}                         & 91.09\scriptsize{±1.60}            & \underline{78.57\scriptsize{±0.12}}    & \underline{66.72\scriptsize{±0.07}}    \\
CORES \citep{cheng2020learning}                 & 94.16\scriptsize{±0.11} & 94.45\scriptsize{±0.14} & 94.88\scriptsize{±0.31} & 94.74\scriptsize{±0.03} & 95.25\scriptsize{±0.09}                         & 91.66\scriptsize{±0.09}            & 73.87\scriptsize{±0.16}     & 55.72\scriptsize{±0.42}    \\
SOP  \citep{sopliu22w}                                         & \textbf{96.38\scriptsize{±0.31}} & \underline{95.28\scriptsize{±0.13}} & \underline{95.31\scriptsize{±0.10}} & \underline{95.39\scriptsize{±0.11}} & \underline{95.61\scriptsize{±0.13}} & \underline{93.24\scriptsize{±0.21}} & \textbf{78.91\scriptsize{±0.43}} & \underline{67.81\scriptsize{±0.23}} \\ \midrule
Ours                  & \underline{96.21\scriptsize{±0.29}} & \textbf{96.06\scriptsize{±0.07}} & \textbf{95.98\scriptsize{±0.12}} & \textbf{96.10\scriptsize{±0.05}} & \textbf{96.40\scriptsize{±0.03}}  & \textbf{93.55\scriptsize{±0.14}} & 78.53\scriptsize{±0.21}     & \textbf{68.07\scriptsize{±0.33}}    \\ \bottomrule
\end{tabular}%
}
\end{table}

\begin{table}[h!]
\centering
\caption{Test accuracy comparison of realistic noisy labels on Clothing1M and WebVision for \textbf{noisy label learning}. 
The best results are indicated in \textbf{bold} and the second best results are indicated in \underline{underline}. Our results are averaged over 3 independent runs. For Clothing1M, we use ImageNet-1K pre trained ResNet50 as the backbone. For WebVision, InceptionResNetv2 is used as the backbone.}
\label{tab:append-noisy-web}
\resizebox{0.4 \textwidth}{!}{%
\begin{tabular}{@{}l|cc@{}}
\toprule
Dataset     & Clothing1M           & WebVision           \\ \midrule
CE        & 69.10                & -                   \\
Forward  \citep{Patrini2016MakingDN}     & 69.80                & 61.10               \\
MentorNet \citep{jiang2018mentornet} & 66.17 & 63.00 \\ 
Co-Teaching \citep{Han2018CoteachingRT} & 69.20                & 63.60               \\
DivideMix \citep{Li2020DivideMixLW}   & \textbf{74.76}               & \underline{77.32}               \\
ELR \citep{Liu2020EarlyLearningRP}        & 72.90               & 76.20               \\
CORES \citep{cheng2020learning}       & 73.20                & -                   \\
SOP  \citep{sopliu22w}       & 73.50                & 76.60               \\ \midrule
Ours        & \underline{74.02\scriptsize{±0.12}} & \textbf{79.37\scriptsize{±0.09}} \\ \bottomrule
\end{tabular}%
}
\end{table}


\subsection{Mixed Imprecise Label Learning}

\subsubsection{Setup}

To create a mixture of various imprecise label configurations, we select CIFAR-10 and CIFAR-100 as base datasets. 
We first uniformly sample $l / C$ labeled samples from each class to form the labeled dataset and treat the remaining samples as the unlabeled dataset.
Based on the labeled dataset, we generate partially labeled datasets by flipping negative labels to false positive labels with the partial ratio $q$. 
After obtaining the partial labels, we randomly select $\eta$ percentage of samples from each class, and recreate the partial labels for them by flipping the ground truth label uniformly to another class. 
In this setting, unlabeled data, partially labeled data, and noisy labeled data exist simultaneously, which is very challenging and more closely resembles realistic situations. 
For CIFAR-10, we set $l \in \{1000, 5000, 50000\}$, and for CIFAR-100, we set $l \in \{5000, 10000, 50000\}$. 
Similarly in the partial label setting, we set $q \in \{0.1, 0.3, 0.5\}$ for CIFAR-10, and $q \in \{0.01, 0.05, 0.1\}$ for CIFAR-100.
For noisy labels, we set $\eta \in \{0.1, 0.2, 0.3\}$ for both datasets.


\subsubsection{Results}

We provide a more complete version of \cref{tab:main-mix} in \cref{tab:append-partial-noisy}. 
On partial noisy labels of CIFAR-10 with partial ratio $0.5$ and of CIFAR-100 with partial ratio $0.1$, most baseline methods are more robust or even fail to perform. 
However, our ILL still shows very robust performance with minor performance degradation as increase of noise ratios.

\begin{table}[h!]
\centering
\caption{Test accuracy comparison of \textbf{mixture of different imprecise labels}. We report results of full labels, partial ratio $q$ of $\{0.1, 0.3, 0.5\}$ for CIFAR-10 and $\{0.01, 0.05, 0.1\}$ for CIFAR-100, and noise ratio $\eta$ of $\{0.1, 0.2, 0.3\}$ for CIFAR-10 and CIFAR-100. 
}
\label{tab:append-partial-noisy}
\resizebox{\textwidth}{!}{%
\begin{tabular}{@{}c|c|c|l|cccc@{}}
\toprule
Dataset &
  \# Labels &
  Partial Ratio $q$ &
  \multicolumn{1}{c|}{Noise Ratio $\eta$}  &
  0 &
  0.1 &
  0.2 &
  0.3 \\ \midrule
 &
   &
   &
  PiCO+ \citep{wang2022pico+} &
  95.99\scriptsize{±0.03} &
  93.64 &
  93.13 &
  92.18 \\
 &
   &
   &
  IRNet \citep{lian2022irnet} &
  - &
  93.44 &
  92.57 &
  92.38 \\
 &
   &
   &
  DALI \citep{xu2023dali} &
  - &
  94.15 &
  94.04 &
  93.77 \\
 &
   &
   &
  PiCO+ w/ Mixup \citep{xu2023dali} &
  - &
  94.58 &
  94.74 &
  94.43 \\
 &
   &
   &
  DALI w/ Mixup \citep{xu2023dali} &
  - &
  95.83 &
  95.86 &
  95.75 \\
 &
   &
  \multirow{-6}{*}{0.1} &
  Ours &
  \textbf{96.55\scriptsize{±0.08}} &
  \textbf{96.47\scriptsize{±0.11}} &
  \textbf{96.09\scriptsize{±0.20}} &
  \textbf{95.83\scriptsize{±0.05}} \\ \cmidrule(l){3-8} 
 &
   &
   &
  PiCO+ \citep{wang2022pico+} &
  95.73\scriptsize{±0.10} &
  92.32 &
  92.22 &
  89.95 \\
 &
   &
   &
  IRNet \citep{lian2022irnet} &
  - &
  92.81 &
  92.18 &
  91.35 \\
 &
   &
   &
  DALI \citep{xu2023dali} &
  - &
  93.44 &
  93.25 &
  92.42 \\
 &
   &
   &
  PiCO+ w/ Mixup \citep{xu2023dali} &
  - &
  94.02 &
  94.03 &
  92.94 \\
 &
   &
   &
  DALI w/ Mixup \citep{xu2023dali} &
  - &
  95.52 &
  95.41 &
  94.67 \\
 &
   &
  \multirow{-6}{*}{0.3} &
  Ours &
  \textbf{96.52\scriptsize{±0.12}} &
  \textbf{96.2\scriptsize{±0.02}} &
  \textbf{95.87\scriptsize{±0.14}} &
  \textbf{95.22\scriptsize{±0.06}} \\ \cmidrule(l){3-8} 
 &
   &
   &
  PiCO+ \citep{wang2022pico+} &
  95.33\scriptsize{±0.06} &
  91.07 &
  89.68 &
  84.08 \\
 &
   &
   &
  IRNet \citep{lian2022irnet} &
  - &
  91.51 &
  90.76 &
  86.19 \\
 &
   &
   &
  DALI \citep{xu2023dali} &
  - &
  92.67 &
  91.83 &
  89.8 \\
 &
   &
   &
   PiCO+  w/ Mixup \citep{xu2023dali} &
  - &
  93.56 &
  92.65 &
  88.21 \\
 &
   &
   &
  DALI w/ Mixup \citep{xu2023dali} &
  - &
  95.19 &
  93.89 &
  92.26 \\
\multirow{-18}{*}{CIFAR-10} &
  \multirow{-18}{*}{50,000} &
  \multirow{-6}{*}{0.5} &
  Ours &
  \textbf{96.28\scriptsize{±0.13}} &
  \textbf{95.82\scriptsize{±0.07}} &
  \textbf{95.28\scriptsize{±0.08}} &
  \textbf{94.35\scriptsize{±0.08}} \\ \bottomrule

   \\ \toprule
   &
   &
   &
  PiCO+ \citep{wang2022pico+} &
  76.29\scriptsize{±0.42} &
  71.42 &
  70.22 &
  66.14 \\
 &
   &
   &
  IRNet \citep{lian2022irnet} &
  - &
  71.17 &
  70.10 &
  68.77 \\
 &
   &
   &
  DALI \citep{xu2023dali} &
  - &
  72.26 &
  71.98 &
  71.04 \\
 &
   &
   &
  PiCO+ w/ Mixup \citep{xu2023dali} &
  - &
  75.04 &
  74.31 &
  71.79 \\
 &
   &
   &
  DALI w/ Mixup \citep{xu2023dali} &
  - &
  76.52 &
  76.55 &
  76.09 \\
 &
   &
  \multirow{-6}{*}{0.01} &
  Ours &
  \textbf{78.08\scriptsize{±0.26}} &
  \textbf{77.53\scriptsize{±0.24}} &
  \textbf{76.96\scriptsize{±0.02}} &
  \textbf{76.43\scriptsize{±0.27}} \\ \cmidrule(l){3-8} 
 &
   &
   &
  PiCO+ \citep{wang2022pico+} &
  76.17\scriptsize{±0.18} &
  69.40 &
  66.67 &
  62.24 \\
 &
   &
   &
  IRNet \citep{lian2022irnet} &
  - &
  70.73 &
  69.33 &
  68.09 \\
 &
   &
   &
  DALI \citep{xu2023dali} &
  - &
  72.28 &
  71.35 &
  70.05 \\
 &
   &
   &
   PiCO+ w/ Mixup \citep{xu2023dali} &
  - &
  73.06 &
  71.37 &
  67.56 \\
 &
   &
   &
 DALI w/ Mixup \citep{xu2023dali} &
  - &
  76.87 &
  75.23 &
  74.49 \\
 &
   &
  \multirow{-6}{*}{0.05} &
  Ours &
  \textbf{76.95\scriptsize{±0.46}} &
  \textbf{77.07\scriptsize{±0.16}} &
  \textbf{76.34\scriptsize{±0.08}} &
  \textbf{75.13\scriptsize{±0.63}} \\ \cmidrule(l){3-8} 
 &
   &
   &
  PiCO+ \citep{wang2022pico+} &
  75.55\scriptsize{±0.21} &
  - &
  - &
  - \\
 &
   &
   &
  IRNet \citep{lian2022irnet} &
  - &
  - &
  - &
  - \\
 &
   &
   &
  DALI \citep{xu2023dali} &
  - &
  - &
  - &
  - \\
 &
   &
   &
  PiCO+ w/ Mixup \citep{xu2023dali} &
  - &
  - &
  - &
  - \\
 &
   &
   &
  DALI w/ Mixup \citep{xu2023dali} &
  - &
  - &
  - &
  - \\
\multirow{-18}{*}{CIFAR-100} &
  \multirow{-18}{*}{50,000} &
  \multirow{-6}{*}{0.1} &
  Ours &
  \textbf{76.41\scriptsize{±1.02}} &
  \textbf{75.50\scriptsize{±0.54}} &
  \textbf{74.67\scriptsize{±0.30}} &
  \textbf{73.88\scriptsize{±0.60}} \\ 
  \bottomrule

\end{tabular}%
}
\end{table}

\subsection{Ablation on Strong-Augmentation and Entropy Loss}

We provide the ablation study on the strong-augmentation and entropy loss components here, which are common techniques in each setting \citep{sohn2020fixmatch,wang2022pico,sopliu22w}. 
For example, in SSL, strong-weak augmentation is an important strategy for SSL algorithms widely used in existing works such as FixMatch \citep{sohn2020fixmatch} and FlexMatch \citep{zhang2021flexmatch}.
Thus, it is important to adopt strong-weak augmentation to achieve better performance in SSL \citep{wang2023freematch,chen2023softmatch,usb2022}.
This is similar in PLL settings \citep{wang2022pico,revisitpllwu22l}. 
PiCO \citep{wang2022pico,revisitpllwu22l} also used strong augmentation).
Strong-weak augmentation and entropy loss are also adopted in SOP \citep{sopliu22w} of NLL. 
However, we found these techniques are less important for our formulation of NLL. 
We provide an ablation study on the entropy loss of SSL, and both techniques for NLL and PLL here to demonstrate our discussions. 

\begin{table}[h]
    \begin{minipage}{.24\linewidth}
      \caption{SSL ablation}
      \vspace{0.1in}
      \centering
        \centering
        \resizebox{\textwidth}{!}{%
        \begin{tabular}{c|cc}
        \hline
                                                                            & \begin{tabular}[c]{@{}c@{}}CIFAR100 \\ $l$=200 \end{tabular} & \begin{tabular}[c]{@{}c@{}}STL10\\ $l$=40\end{tabular} \\ \hline
        Ours                                                                & 22.06                                                          & 11.09                                                     \\ \hline
        \begin{tabular}[c]{@{}c@{}}Ours\\  w/o \\ ent. \end{tabular} & 22.41                                                          & 11.23                                                     \\ \hline
        \end{tabular}%
        }
    \end{minipage}%
    \hfill
    \begin{minipage}{.24\linewidth}
      \centering
        \caption{PLL ablation}
        \vspace{0.1in}
            \resizebox{\textwidth}{!}{%
            \begin{tabular}{c|cc}
            \hline
            \multicolumn{1}{c|}{}                                      & \begin{tabular}[c]{@{}c@{}}CIFAR10\\ $q=0.5$\end{tabular} & \begin{tabular}[c]{@{}c@{}}CIFAR100\\ $q=0.1$\end{tabular} \\ \hline
            \multicolumn{1}{c|}{PiCO}                                  & 93.58                                                     & 69.91                                                      \\ \hline
            \multicolumn{1}{c|}{Ours}                                  & 95.91                                                     & 74.00                                                      \\ \hline
            \begin{tabular}[c]{@{}c@{}}PiCO\\ w/o\\ s. a.\end{tabular} & 91.78                                                     & 66.43                                                      \\ \hline
            \begin{tabular}[c]{@{}c@{}}Ours\\ w/o\\ s. a.\end{tabular} & 94.53                                                     & 72.69                                                      \\ \hline
            \begin{tabular}[c]{@{}c@{}}Ours\\ w/o\\ ent.\end{tabular}  & 95.87                                                     & 73.75                                                      \\ \hline
            \end{tabular}%
            }
    \end{minipage} 
    \hfill
    \begin{minipage}{.24\linewidth}
      \centering
        \caption{NLL ablation}
        \vspace{0.1in}
            \resizebox{\textwidth}{!}{%
            \begin{tabular}{c|cc}
            \hline
                                                                       & \begin{tabular}[c]{@{}c@{}}CIFAR10\\ $\eta=0.5$\end{tabular} & \begin{tabular}[c]{@{}c@{}}CIFAR100\\ $\eta=0.1$\end{tabular} \\ \hline
            SOP                                                        & 94.00                                                        & 63.30                                                         \\ \hline
            Ours                                                       & 94.31                                                        & 66.46                                                         \\ \hline
            \begin{tabular}[c]{@{}c@{}}SOP\\ w/o\\ s. a.\end{tabular}  & 66.85                                                        & 36.60                                                         \\ \hline
            \begin{tabular}[c]{@{}c@{}}Ours\\ w/o\\ s. a.\end{tabular} & 93.56                                                        & 65.89                                                         \\ \hline
            \begin{tabular}[c]{@{}c@{}}SOP\\ w/o\\ ent.\end{tabular}   & 93.04                                                        & 62.85                                                         \\ \hline
            \begin{tabular}[c]{@{}c@{}}Ours\\ w/o\\ ent.\end{tabular}  & 94.16                                                        & 66.12                                                         \\ \hline
            \end{tabular}%
            }
    \end{minipage} 
    \hfill
    \begin{minipage}{.24\linewidth}
      \centering
        \caption{Runtime Analysis on CIFAR-100}
        \vspace{0.1in}
            \resizebox{\textwidth}{!}{%
                \begin{tabular}{@{}ccc@{}}
                \toprule
                Setting & Algorithm & CIFAR-100 Avg. \\ & &  Runtime (s/iter) \\ \midrule
                SSL     & FreeMatch & 0.2157                          \\
                SSL     & Ours      & 0.1146                          \\ \midrule
                PLL     & PiCO      & 0.3249                          \\
                PLL     & Ours      & 0.2919                          \\ \midrule
                NLL     & SOP       & 0.1176                          \\
                NLL     & Ours       & 0.1021                          \\ \bottomrule
                \end{tabular}%
            }
    \label{tab:runtime}
    \end{minipage} 
\end{table}

\subsection{Runtime Analysis}

We provide the runtime analysis on CIFAR-100 of our method on different settings, compared with the SOTA baselines. 
We compute the average runtime from all training iterations on CIFAR-100. 
The results are shown in \cref{tab:runtime}. 
Our method in general present faster runtime without complex design such as contrastive loss.

\newpage


\clearpage
\newpage
\section*{NeurIPS Paper Checklist}

\begin{enumerate}

\item {\bf Claims}
    \item[] Question: Do the main claims made in the abstract and introduction accurately reflect the paper's contributions and scope?
    \item[] Answer: \answerYes{} 
    \item[] Justification: We discussed our contribution in introduction.
    \item[] Guidelines: 
    \begin{itemize}
        \item The answer NA means that the abstract and introduction do not include the claims made in the paper.
        \item The abstract and/or introduction should clearly state the claims made, including the contributions made in the paper and important assumptions and limitations. A No or NA answer to this question will not be perceived well by the reviewers. 
        \item The claims made should match theoretical and experimental results, and reflect how much the results can be expected to generalize to other settings. 
        \item It is fine to include aspirational goals as motivation as long as it is clear that these goals are not attained by the paper. 
    \end{itemize}

\item {\bf Limitations}
    \item[] Question: Does the paper discuss the limitations of the work performed by the authors?
    \item[] Answer: \answerYes{} 
    \item[] Justification: We discussed our limitation in conclusion.
    \item[] Guidelines:
    \begin{itemize}
        \item The answer NA means that the paper has no limitation while the answer No means that the paper has limitations, but those are not discussed in the paper. 
        \item The authors are encouraged to create a separate "Limitations" section in their paper.
        \item The paper should point out any strong assumptions and how robust the results are to violations of these assumptions (e.g., independence assumptions, noiseless settings, model well-specification, asymptotic approximations only holding locally). The authors should reflect on how these assumptions might be violated in practice and what the implications would be.
        \item The authors should reflect on the scope of the claims made, e.g., if the approach was only tested on a few datasets or with a few runs. In general, empirical results often depend on implicit assumptions, which should be articulated.
        \item The authors should reflect on the factors that influence the performance of the approach. For example, a facial recognition algorithm may perform poorly when image resolution is low or images are taken in low lighting. Or a speech-to-text system might not be used reliably to provide closed captions for online lectures because it fails to handle technical jargon.
        \item The authors should discuss the computational efficiency of the proposed algorithms and how they scale with dataset size.
        \item If applicable, the authors should discuss possible limitations of their approach to address problems of privacy and fairness.
        \item While the authors might fear that complete honesty about limitations might be used by reviewers as grounds for rejection, a worse outcome might be that reviewers discover limitations that aren't acknowledged in the paper. The authors should use their best judgment and recognize that individual actions in favor of transparency play an important role in developing norms that preserve the integrity of the community. Reviewers will be specifically instructed to not penalize honesty concerning limitations.
    \end{itemize}

\item {\bf Theory Assumptions and Proofs}
    \item[] Question: For each theoretical result, does the paper provide the full set of assumptions and a complete (and correct) proof?
    \item[] Answer: \answerYes{} 
    \item[] Justification: All are stated in Appendix.
    \item[] Guidelines:
    \begin{itemize}
        \item The answer NA means that the paper does not include theoretical results. 
        \item All the theorems, formulas, and proofs in the paper should be numbered and cross-referenced.
        \item All assumptions should be clearly stated or referenced in the statement of any theorems.
        \item The proofs can either appear in the main paper or the supplemental material, but if they appear in the supplemental material, the authors are encouraged to provide a short proof sketch to provide intuition. 
        \item Inversely, any informal proof provided in the core of the paper should be complemented by formal proofs provided in appendix or supplemental material.
        \item Theorems and Lemmas that the proof relies upon should be properly referenced. 
    \end{itemize}

    \item {\bf Experimental Result Reproducibility}
    \item[] Question: Does the paper fully disclose all the information needed to reproduce the main experimental results of the paper to the extent that it affects the main claims and/or conclusions of the paper (regardless of whether the code and data are provided or not)?
    \item[] Answer: \answerYes{}
    \item[] Justification: We present all details in both main paper and Appendix.
    \item[] Guidelines:
    \begin{itemize}
        \item The answer NA means that the paper does not include experiments.
        \item If the paper includes experiments, a No answer to this question will not be perceived well by the reviewers: Making the paper reproducible is important, regardless of whether the code and data are provided or not.
        \item If the contribution is a dataset and/or model, the authors should describe the steps taken to make their results reproducible or verifiable. 
        \item Depending on the contribution, reproducibility can be accomplished in various ways. For example, if the contribution is a novel architecture, describing the architecture fully might suffice, or if the contribution is a specific model and empirical evaluation, it may be necessary to either make it possible for others to replicate the model with the same dataset, or provide access to the model. In general. releasing code and data is often one good way to accomplish this, but reproducibility can also be provided via detailed instructions for how to replicate the results, access to a hosted model (e.g., in the case of a large language model), releasing of a model checkpoint, or other means that are appropriate to the research performed.
        \item While NeurIPS does not require releasing code, the conference does require all submissions to provide some reasonable avenue for reproducibility, which may depend on the nature of the contribution. For example
        \begin{enumerate}
            \item If the contribution is primarily a new algorithm, the paper should make it clear how to reproduce that algorithm.
            \item If the contribution is primarily a new model architecture, the paper should describe the architecture clearly and fully.
            \item If the contribution is a new model (e.g., a large language model), then there should either be a way to access this model for reproducing the results or a way to reproduce the model (e.g., with an open-source dataset or instructions for how to construct the dataset).
            \item We recognize that reproducibility may be tricky in some cases, in which case authors are welcome to describe the particular way they provide for reproducibility. In the case of closed-source models, it may be that access to the model is limited in some way (e.g., to registered users), but it should be possible for other researchers to have some path to reproducing or verifying the results.
        \end{enumerate}
    \end{itemize}

\item {\bf Open access to data and code}
    \item[] Question: Does the paper provide open access to the data and code, with sufficient instructions to faithfully reproduce the main experimental results, as described in supplemental material?
    \item[] Answer: \answerYes{} 
    \item[] Justification: We use publically available data and code will be released.
    \item[] Guidelines:
    \begin{itemize}
        \item The answer NA means that paper does not include experiments requiring code.
        \item Please see the NeurIPS code and data submission guidelines (\url{https://nips.cc/public/guides/CodeSubmissionPolicy}) for more details.
        \item While we encourage the release of code and data, we understand that this might not be possible, so “No” is an acceptable answer. Papers cannot be rejected simply for not including code, unless this is central to the contribution (e.g., for a new open-source benchmark).
        \item The instructions should contain the exact command and environment needed to run to reproduce the results. See the NeurIPS code and data submission guidelines (\url{https://nips.cc/public/guides/CodeSubmissionPolicy}) for more details.
        \item The authors should provide instructions on data access and preparation, including how to access the raw data, preprocessed data, intermediate data, and generated data, etc.
        \item The authors should provide scripts to reproduce all experimental results for the new proposed method and baselines. If only a subset of experiments are reproducible, they should state which ones are omitted from the script and why.
        \item At submission time, to preserve anonymity, the authors should release anonymized versions (if applicable).
        \item Providing as much information as possible in supplemental material (appended to the paper) is recommended, but including URLs to data and code is permitted.
    \end{itemize}

\item {\bf Experimental Setting/Details}
    \item[] Question: Does the paper specify all the training and test details (e.g., data splits, hyperparameters, how they were chosen, type of optimizer, etc.) necessary to understand the results?
    \item[] Answer: \answerYes{} 
    \item[] Justification: All results are obtained with 3 runs using different seeds and error bars are reported.
    \item[] Guidelines:
    \begin{itemize}
        \item The answer NA means that the paper does not include experiments.
        \item The experimental setting should be presented in the core of the paper to a level of detail that is necessary to appreciate the results and make sense of them.
        \item The full details can be provided either with the code, in appendix, or as supplemental material.
    \end{itemize}

\item {\bf Experiment Statistical Significance}
    \item[] Question: Does the paper report error bars suitably and correctly defined or other appropriate information about the statistical significance of the experiments?
    \item[] Answer: \answerYes{} 
    \item[] Justification: We report the standard deviation in the results, which are averaged over 3 independent runs across different experiments. 
    \item[] Guidelines:
    \begin{itemize}
        \item The answer NA means that the paper does not include experiments.
        \item The authors should answer "Yes" if the results are accompanied by error bars, confidence intervals, or statistical significance tests, at least for the experiments that support the main claims of the paper.
        \item The factors of variability that the error bars are capturing should be clearly stated (for example, train/test split, initialization, random drawing of some parameter, or overall run with given experimental conditions).
        \item The method for calculating the error bars should be explained (closed form formula, call to a library function, bootstrap, etc.)
        \item The assumptions made should be given (e.g., Normally distributed errors).
        \item It should be clear whether the error bar is the standard deviation or the standard error of the mean.
        \item It is OK to report 1-sigma error bars, but one should state it. The authors should preferably report a 2-sigma error bar than state that they have a 96\% CI, if the hypothesis of Normality of errors is not verified.
        \item For asymmetric distributions, the authors should be careful not to show in tables or figures symmetric error bars that would yield results that are out of range (e.g. negative error rates).
        \item If error bars are reported in tables or plots, The authors should explain in the text how they were calculated and reference the corresponding figures or tables in the text.
    \end{itemize}

\item {\bf Experiments Compute Resources}
    \item[] Question: For each experiment, does the paper provide sufficient information on the computer resources (type of compute workers, memory, time of execution) needed to reproduce the experiments?
    \item[] Answer: \answerYes{} 
    \item[] Justification: We showed in experiments.
    \item[] Guidelines:
    \begin{itemize}
        \item The answer NA means that the paper does not include experiments.
        \item The paper should indicate the type of compute workers CPU or GPU, internal cluster, or cloud provider, including relevant memory and storage.
        \item The paper should provide the amount of compute required for each of the individual experimental runs as well as estimate the total compute. 
        \item The paper should disclose whether the full research project required more compute than the experiments reported in the paper (e.g., preliminary or failed experiments that didn't make it into the paper). 
    \end{itemize}
    
\item {\bf Code Of Ethics}
    \item[] Question: Does the research conducted in the paper conform, in every respect, with the NeurIPS Code of Ethics \url{https://neurips.cc/public/EthicsGuidelines}?
    \item[] Answer: \answerYes{} 
    \item[] Justification: All behaviors follows NeurIPS Code of Ethics.
    \item[] Guidelines:
    \begin{itemize}
        \item The answer NA means that the authors have not reviewed the NeurIPS Code of Ethics.
        \item If the authors answer No, they should explain the special circumstances that require a deviation from the Code of Ethics.
        \item The authors should make sure to preserve anonymity (e.g., if there is a special consideration due to laws or regulations in their jurisdiction).
    \end{itemize}

\item {\bf Broader Impacts}
    \item[] Question: Does the paper discuss both potential positive societal impacts and negative societal impacts of the work performed?
    \item[] Answer: \answerYes{} 
    \item[] Justification: We discussed in conclusion.
    \item[] Guidelines:
    \begin{itemize}
        \item The answer NA means that there is no societal impact of the work performed.
        \item If the authors answer NA or No, they should explain why their work has no societal impact or why the paper does not address societal impact.
        \item Examples of negative societal impacts include potential malicious or unintended uses (e.g., disinformation, generating fake profiles, surveillance), fairness considerations (e.g., deployment of technologies that could make decisions that unfairly impact specific groups), privacy considerations, and security considerations.
        \item The conference expects that many papers will be foundational research and not tied to particular applications, let alone deployments. However, if there is a direct path to any negative applications, the authors should point it out. For example, it is legitimate to point out that an improvement in the quality of generative models could be used to generate deepfakes for disinformation. On the other hand, it is not needed to point out that a generic algorithm for optimizing neural networks could enable people to train models that generate Deepfakes faster.
        \item The authors should consider possible harms that could arise when the technology is being used as intended and functioning correctly, harms that could arise when the technology is being used as intended but gives incorrect results, and harms following from (intentional or unintentional) misuse of the technology.
        \item If there are negative societal impacts, the authors could also discuss possible mitigation strategies (e.g., gated release of models, providing defenses in addition to attacks, mechanisms for monitoring misuse, mechanisms to monitor how a system learns from feedback over time, improving the efficiency and accessibility of ML).
    \end{itemize}
    
\item {\bf Safeguards}
    \item[] Question: Does the paper describe safeguards that have been put in place for responsible release of data or models that have a high risk for misuse (e.g., pretrained language models, image generators, or scraped datasets)?
    \item[] Answer: \answerNA{} 
    \item[] Justification: NA
    \item[] Guidelines:
    \begin{itemize}
        \item The answer NA means that the paper poses no such risks.
        \item Released models that have a high risk for misuse or dual-use should be released with necessary safeguards to allow for controlled use of the model, for example by requiring that users adhere to usage guidelines or restrictions to access the model or implementing safety filters. 
        \item Datasets that have been scraped from the Internet could pose safety risks. The authors should describe how they avoided releasing unsafe images.
        \item We recognize that providing effective safeguards is challenging, and many papers do not require this, but we encourage authors to take this into account and make a best faith effort.
    \end{itemize}

\item {\bf Licenses for existing assets}
    \item[] Question: Are the creators or original owners of assets (e.g., code, data, models), used in the paper, properly credited and are the license and terms of use explicitly mentioned and properly respected?
    \item[] Answer:\answerYes{}
    \item[] Justification: We discussed in experiments.
    \item[] Guidelines:
    \begin{itemize}
        \item The answer NA means that the paper does not use existing assets.
        \item The authors should cite the original paper that produced the code package or dataset.
        \item The authors should state which version of the asset is used and, if possible, include a URL.
        \item The name of the license (e.g., CC-BY 4.0) should be included for each asset.
        \item For scraped data from a particular source (e.g., website), the copyright and terms of service of that source should be provided.
        \item If assets are released, the license, copyright information, and terms of use in the package should be provided. For popular datasets, \url{paperswithcode.com/datasets} has curated licenses for some datasets. Their licensing guide can help determine the license of a dataset.
        \item For existing datasets that are re-packaged, both the original license and the license of the derived asset (if it has changed) should be provided.
        \item If this information is not available online, the authors are encouraged to reach out to the asset's creators.
    \end{itemize}

\item {\bf New Assets}
    \item[] Question: Are new assets introduced in the paper well documented and is the documentation provided alongside the assets?
    \item[] Answer: \answerNA{} 
    \item[] Justification: NA
    \item[] Guidelines:
    \begin{itemize}
        \item The answer NA means that the paper does not release new assets.
        \item Researchers should communicate the details of the dataset/code/model as part of their submissions via structured templates. This includes details about training, license, limitations, etc. 
        \item The paper should discuss whether and how consent was obtained from people whose asset is used.
        \item At submission time, remember to anonymize your assets (if applicable). You can either create an anonymized URL or include an anonymized zip file.
    \end{itemize}

\item {\bf Crowdsourcing and Research with Human Subjects}
    \item[] Question: For crowdsourcing experiments and research with human subjects, does the paper include the full text of instructions given to participants and screenshots, if applicable, as well as details about compensation (if any)? 
    \item[] Answer: \answerNA{} 
    \item[] Justification: NA.
    \item[] Guidelines:
    \begin{itemize}
        \item The answer NA means that the paper does not involve crowdsourcing nor research with human subjects.
        \item Including this information in the supplemental material is fine, but if the main contribution of the paper involves human subjects, then as much detail as possible should be included in the main paper. 
        \item According to the NeurIPS Code of Ethics, workers involved in data collection, curation, or other labor should be paid at least the minimum wage in the country of the data collector. 
    \end{itemize}

\item {\bf Institutional Review Board (IRB) Approvals or Equivalent for Research with Human Subjects}
    \item[] Question: Does the paper describe potential risks incurred by study participants, whether such risks were disclosed to the subjects, and whether Institutional Review Board (IRB) approvals (or an equivalent approval/review based on the requirements of your country or institution) were obtained?
    \item[] Answer: \answerNA{} 
    \item[] Justification: NA.
    \item[] Guidelines:
    \begin{itemize}
        \item The answer NA means that the paper does not involve crowdsourcing nor research with human subjects.
        \item Depending on the country in which research is conducted, IRB approval (or equivalent) may be required for any human subjects research. If you obtained IRB approval, you should clearly state this in the paper. 
        \item We recognize that the procedures for this may vary significantly between institutions and locations, and we expect authors to adhere to the NeurIPS Code of Ethics and the guidelines for their institution. 
        \item For initial submissions, do not include any information that would break anonymity (if applicable), such as the institution conducting the review.
    \end{itemize}

\end{enumerate}

\end{document}